\documentclass[letterpaper]{article}

\usepackage{arxiv}

\usepackage[utf8]{inputenc} 
\usepackage[T1]{fontenc}    
\usepackage{hyperref}       
\usepackage{url}            
\usepackage{booktabs}       
\usepackage{amsfonts}       
\usepackage{microtype}      
\usepackage{amsmath}
\usepackage{graphicx}
\usepackage{algorithm}
\usepackage[noend]{algpseudocode}
\usepackage{tikz}
\usetikzlibrary{shapes,arrows}

\usepackage[style=ieee,doi=false,isbn=false,url=false]{biblatex}
\title{Path Planning in Dynamic Environments Using Time-Warped Grids and a Parallel Implementation}

\author{
  Siavash Farzan \\
  Institute for Robotics and Intelligent Machines\\
  Georgia Institute of Technology\\
  Atlanta, GA 30332 \\
  \texttt{sfarzan@gatech.edu}
   \And
 Guilherme N. DeSouza \\
  ViGIR Lab, EECS Department \\ 
  University of Missouri\\
  Columbia, MO 65211 \\
  \texttt{DeSouzaG@missouri.edu} \\
}

\date{}

\begin{document}
\maketitle

\begin{abstract}
This paper proposes a solution to the problem of smooth path planning
for mobile robots in dynamic and unknown environments. A novel concept
of Time-Warped Grid is introduced to predict the pose of obstacles
in the environment and avoid collisions. The algorithm is implemented
using C/C++ and the CUDA programming environment, and combines stochastic
estimation (Kalman filter), harmonic potential fields and a rubber
band model, and it translates naturally into the parallel paradigm
of GPU programming. In simple terms, time-warped grids are progressively
wider orbits around the mobile robot. Those orbits represent the variable
time intervals estimated by the robot to reach detected obstacles.

The proposed method was tested using several simulation scenarios
for the Pioneer P3-DX robot, which demonstrated the robustness of
the algorithm by finding the optimum path in terms of smoothness,
distance, and collision-free, in both static or dynamic environments, and
with large number of obstacles.
\end{abstract}


\section{Introduction}
The problem of path planning of mobile robots consists of the trajectory
planning of a mobile robot from a start state to a final state. The
objective of such start-to-goal path planning is to find the best
suitable path between two points while avoiding collisions with both static
and moving obstacles. In this case, the suitability of the ``best''
path is determined by a function representing the occupancy of the
space, while the selected robot trajectory corresponds to the optimal
path on the surface of this occupancy function according to some criteria (e.g. smoothness, length, etc.). 

In such problems, the trade off between performing quick, shallow
searches for sub-optimal solutions and performing deep searches for
the price of heavy computation has always been a practical and important issue.
The application of parallel algorithms however, can minimize or even
eliminate this issue by enabling deeper searches without losing accuracy
or real-time applicability. In addition to saving time, parallel computation
can lead to extra advantages \cite{Barney10} such as solving the same problems at
larger scales, providing redundancy of control, overcoming memory constraints, and performing
remote processing (e.g. tele-operation of robots). 

However, these advantages do not come with a low price tag. For effective
use of parallel computing, the computational problem must be partitioned
into discrete parts, or tasks, to be executed simultaneously.
This partitioning into discrete parts is carried out by identifying
independent and self-contained sub tasks where the overall result
can be readily extracted from the sum of the individual results. As
it will be explained in further details later, this can be a daunting operation.

In this paper, a solution to the problem of smooth path planning for
mobile robots in dynamic and unknown environments is proposed. A novel
concept of time-warped grid is introduced to predict the pose of obstacles
in the environment while avoiding collisions with those obstacles. The
concept combines stochastic estimation (Kalman filter), harmonic potential
fields and a rubber band model, and it translates naturally into the
parallel paradigm of GPU programming. 
The intuition behind the proposed method for dealing with dynamic environments is derived from the analogy of pedestrians ``jaywalking'' or crossing streets on which vehicles are driving. When jaywalking, pedestrians anticipate where the vehicles will be at future times and adjust their path accordingly. If a person were to consider only a snapshot of the street and assume a fixed position for each vehicle, then he/she would be in a significant risk of being run over if tried to cross the street. Similarly in this work, the robot predicts the future configuration of moving obstacles by using a combination of Kalman filter and time-warped grids. Consequently, a path is plotted for the future positions of the obstacles, rather than the current ones. 



In simple terms, time-warped grids are progressively wider orbits around the mobile robot. Those orbits represent the variable time intervals estimated by the robot to reach detected obstacles. The main idea of time-warped grid is to acknowledge that the further the obstacle is, the more delayed is its impact on the path. The use of time-warped grid allows the
system to address at the same time, the problems of convergence, speed,
and moving obstacles in the calculation of a smooth path for the mobile
robot without any prior knowledge regarding the environment. All assumptions
made by the system derive from a laser sensor mounted on the robot
and a localization system (e.g. vision-based landmark localization
\cite{DeSKak02}) that provides distance to the goal, even when it
lies outside the range of the robot sensors in large environments.

As the experiments performed demonstrate, the path obtained by the
predictive aspect of our method is not only short, but it also contains
no loops, no sharp turns, and no changes of speed of the robot, making
it ideal for carrying of delicate materials or for wheelchair navigation.
Our experiments also demonstrate the robustness of the method, which
can always find an optimum path -- i.e in terms of smoothness, distance,
and collision-free -- either in static or dynamic environments, even
with a very large number of obstacles.

\section{Related Work}

Unlike \emph{coverage path planning} where the emphasis is in sweeping
out the space sensed by the robot -- e.g. floor cleaning, lawn mowing,
harvesting, etc. \cite{Cho01}, the objective of \emph{start-goal
path planning} is to find suitable paths between two points while
avoiding collision with static or moving obstacles. The ability to generate an optimal path from an initial point to a final destination in real-time is still one of the key challenges in Robotics. The area is becoming even more influential with the near deployment of self-driving autonomous vehicles \cite{Bast16}.

Different techniques have been proposed in this domain,
where the challenges become increasingly difficult with the size of the environment and the number of moving obstacles. From the early algorithms, the focus often turned into
finding cost-minimal paths through the robot environment,
and the use of maps became so attached to the problem that many confuse start-goal and map-based as synonymous. Indeed, most of the approaches in the literature rely on some sort of map or grid, and the large size of the environment and consequent number of grids rendered many of these methods to off-line use only. Moreover, the complexity and uncertainty of the path planning problem increase greatly in dynamic environments due to the change of the entire information in the environment along with the movement of obstacles.

An early and important algorithm to address the path planning problem in dynamic environments is known as diffusion process over grid-based maps, first presented by Steels \cite{Steels88}, and further developed by Schmidt et al. \cite{SchAza92}. In diffusion method, the environment is discretized into a grid of cells, where the cell representing the goal applies an attracting force in succession to the neighbouring free cells all the way to the cell occupied by the robot. These attractive forces are modeled by fluid diffusion equations, hence the name of the method. The robot proceeds towards the cell with the highest level of force computed by a gradient-based iterative algorithm. The diffusion method can be implemented as a two layer cellular neural network \cite{Sie94}, and use of an unsteady diffusion equation model \cite{SchAza92} makes it applicable to time-varying environments.
More recently, Vázquez-Otero et al. \cite{VazMun12} used reaction-diffusion dynamic models based on biological processes 
with advantages of smoother trajectories and tolerance to noisy data.

Diffusion-based methods allow on-line path planning and result in short and collision-free paths, without suffering from problems with local minima.
However, the algorithm is global and require the prior knowledge of the robot’s environment, thus cannot deal with more advanced navigation tasks such as exploration of unknown environment or multi-robot navigation. Moreover, the diffusion equation parameters are difficult to tune, and the implementation is highly computationally demanding, making the algorithm less favorable for real-time or large environment applications.

Borenstein and Koren \cite{BorKor91} presented an efficient algorithm for path planning called the vector field histogram (VFH) method, further improved by Ulrich and Borenstein \cite{UlrBor98} \cite{UlrBor00} by considering the robot size and choosing a safe and efficient steering direction.
This approach uses a grid-based occupancy mapping, where occupancy information is described by a histogram representation of the robot's environment. The free space and obstacles are localized by their angle as well as their distance relative to the robot. The final steering direction for the robot is selected (based on specified thresholds and proximity to obstacles) from the candidates closest to the goal direction. Due to the statistical nature of the algorithm, the VFH method is very robust to uncertainties in sensor readings and dynamic models, making it ideal for unknown and time-varying environments. While the method is computationally efficient, it does not result in globally optimal paths (since it is a local path planner) and is also prone to dead-end situations.

The dynamic window approach incorporating the motion dynamics of the robot was proposed in \cite{FoxThr97}, and further generalized in \cite{BroKha99}. In this method, the control of the robot is carried out directly in the space of translational and rotational velocities, and the search space for admissible velocities is reduced in three steps over three windows given by: i) approximating trajectories by finite sequences of circular arcs, ii) considering only the next steering command, and iii) restricting the space to the velocities that can be reached within a short time interval. The final path for the robot is generated by maximizing an objective function formed by the intersection of the three windows. Using a dynamic window approach, the acceleration capabilities of the robot are considered, and obstacle avoidance can be performed at high velocities. However, the algorithm is still prone to local minima problems, since it only considers the goal heading without integrating the information about the free space.

Potential field methods, as proposed by \cite{Khatib86}, determine the heading direction of the robot by representing the robot environment as a potential field, where the goal applies an attractive force and obstacles assert repulsive forces to the field. The robot trajectory is calculated by superposition of the two fields and following the low potential along the field according to a fastest descent optimization. Methods based on potential fields are computationally efficient and suitable for path planning with real-time conditions. However, they often fail to find trajectories in congested environments, and can also result in oscillatory motions in narrow passages \cite{KorBor91}. 

Another problem in start-goal planning is regarding the convergence to a solution. In potential fields \cite{HwaAhu92} and the A{*} algorithm \cite{HarRap68}, for example, a guarantee that the system will find a solution either can not be provided at all (potential fields) \cite{Latombe12}, or it can be provided only if the heuristic is guaranteed to always be optimistic (A{*}) -- i.e. the true cost of a path is at least as large as the estimated cost. For the first case, Kim and Khosla \cite{KimKho92} proposed instead a harmonic potential function that eliminates the possibility of local minima in the potential fields, which prevented it from finding a path to the goal. In the latter case of A{*} algorithm, the alternative is to make the heuristic more optimistic, which increases the method's computational complexity, making it less likely to run in real-time. 




As mentioned above, some of these methods require prior knowledge about the environment in a static setting. Violation of this requirement, i.e. existence of moving obstacles, leads to sudden changes and oscillations in the robot path, which can be aggravated by the sensitivity and inaccuracy of the robot sensors. While these consequences may be acceptable for a mobile robot, applications involving autonomous wheelchair navigation can become quite uncomfortable for human passengers. In \cite{HonDes10}, it was proposed a robust method using a rubber band model to smoothen the path and reduce the number of sharp angles obtained from the use of harmonic potential fields alone. While that approach worked well for static environments, it did not address the case of moving obstacles.

In order to address the computational complexity of these methods,
many researchers have recently developed parallel implementation of
path planning algorithms on Graphics Processing Units (GPUs). In \cite{Ble08}, the authors proposed a method for globally optimum path planning using a combination of the A{*} and Dijkstra's algorithms \cite{Dijkstra59}. The two algorithms were modified to take advantage of data parallelism of GPUs, which led to an implementation of edge lists using adjacency tables to reach a remarkable speed-up when compared to traditional C++ implementations.

Also exploiting the nature of these algorithms and the parallel paradigm
of GPU computation, Kider et al. \cite{KidSaf10} proposed a randomized
version of A{*}, called R{*}GPU search. Their main contributions are
certainly the smaller memory requirements when compared to the original
A{*}, the avoidance of local minima by the use of randomly selected
subgoals, and the scalability of the method to high-dimensional planning
problems.


However, even when computational complexity is not an issue, the major drawback of any A{*} based method remains in the difficulty in coping with dynamic environments. That is, A{*} algorithm relies on the \emph{optimism} of the heuristics. Since those heuristics derive from the values of the map cells, potential changes over time in these values or changes in topology of the graphs due to moving obstacles lead to inversions of those heuristics, and hence to loops and/or failure in converging to the goal. Another problem of the A{*} algorithm can be seen in the proposed approach in \cite{Russell16}. 
The main drawback in this approach is that the A{*} uses uniform grid representation which requires large amount of memory for regions that may never be traversed or may not contain any obstacles, affecting the efficiency of the method. This drawback can also happen in the dynamic version of the A{*} algorithm called D{*} \cite{Ste94}, even though it indeed generates optimal trajectories in unknown environments.

In this paper, we address at the same time the problems of convergence,
speed, and highly dynamic environments in the calculation of a smooth path for
a mobile robot without any prior knowledge regarding the
environment (static or dynamic). All assumptions made by the system
derive from a laser sensor mounted on the robot and a localization
system (here assumed to be a vision-based, landmark-based localization \cite{DeSKak02})
which needs to provide only location of the goal, even when it lies outside the
range of the robot sensors in large environments.

\section{Proposed Method}

The proposed method for the path planning problem tackles the limitations
of other systems -- namely the problems of convergence, fast processing, moving
obstacles and sharp paths -- by combining a few concepts. First,
we rely on harmonic potentials to guarantee the calculation of a path
if one exists\footnote{A path may not exist only if obstacles completely block the robot's path
to the goal.}. As it was pointed out earlier, the path produced by harmonic potentials can present
sharp turns, which for many applications, such as autonomous wheelchair
navigation, can produce an uncomfortable experience for the passenger.
So, we re-introduce the idea of a rubber band model \cite{HonDes10}
to smoothen the path created by the harmonic potentials. Since moving obstacles can
also lead to unexpected changes in path, we propose the use of Kalman
filter for stochastic estimation of the positions of the obstacles.
While Kalman filter in itself has been widely used in the past, our main contribution
here is in the combination of Kalman filter and a novel idea of time-warped grid, which will be explained later.

As the experiments performed will demonstrate, the path obtained by the
predictive aspect of our method is not only the shortest possible
path, but it also contains no loops\footnote{Unless a path momentarily does not exist due to a large number of
obstacles (see Section \ref{sec:Experimental-Results}).}, no sharp turns, and no changes of speed of the robot, making it
ideal for carrying of delicate materials or for wheelchair navigation.
Since our method is also relatively computationally intensive, we resort to an efficient parallel implementation using GPUs over the time-warped grids.

\subsection{Harmonic Potential Fields \label{subsec:Potential-Rubber}}

In grid-based maps, the idea is to represent the environment as a
2D grid. Such grid is basically the projection of all objects in the
environment -- in our case detected using a laser range sensor and a vision system --
onto the ground plane. When potential fields are applied on top of
such grids, obstacles are described by high potentials or \emph{hills}
that must be avoided, and start and goal points are the highest and
smallest potentials, respectively. The path towards the destination
is defined along the valleys of the potential field. Unfortunately,
due to interaction between multiple objects, valleys are not unique
in their potential values. Moreover, besides the possibility of multiple
valleys, potential fields can also present local minima. These two conditions can
cause the robot to fail to find a path to the goal. However, these
same problems disappear when we use harmonic potential fields instead
\cite{ConGru93,Robert_08_harmonic}. 

Harmonic functions satisfy the min-max principle, and hence, spontaneous
creation of local minima within the space is impossible. This principle
is satisfied when the Laplace\textquoteright s equation constraint
on the functions is true. In other words, a harmonic function $\phi$
on a domain $\Omega\subset R^{n}$ is a function that satisfies:
\begin{equation*}
\nabla^{2}\phi=\sum_{i=1}^{n}\frac{\partial^{2}\phi}{\partial x_{i}^{2}}=0\textrm{.}
\end{equation*}

This same function can be discretized and the numerical solution of Laplace's equation becomes \cite{ConGru93}:
\begin{equation}
\begin{array}{ccc}
\phi^{(k+1)}(x,y) & =& 
\dfrac{1}{4}[(\phi^{(k)}(x+1,y)+\phi^{(k)}(x-1,y)\quad+ \\ & & \phi^{(k)}(x,y+1)+\phi^{(k)}(x,y-1)]\label{eq:discretelaplace}
\end{array}
\end{equation}
where $\phi^{(k)}(x,\,y)$ represents the discrete sample of $\phi$
at coordinates $( x,\,y )$ of the $\mathbb{R}^{2}$ grid, and $k$ denotes the iteration
number. At each iteration, a grid cell of $\phi$ is updated with
the average value of its neighbors. On a sequential computer, this
solution is usually implemented as follows:
\begin{equation}
\begin{array}{ccc}
\phi^{(k+1)}(x,y) & = & \dfrac{1}{4}[\phi^{(k)}(x+1,y)+\phi^{(k+1)}(x-1,y)\quad+\\
 &  & \phi^{(k)}(x,y+1)+\phi^{(k+1)}(x,y-1)] \textrm{ .} \label{eq:harmonic-potential}
\end{array}
\end{equation}

That is, the next values of the top and left neighbors of the current
cell are updated and used in the calculation of that same cell. This
speed up of the algorithm allows for the values of the next iteration
to quickly propagate through the grid. However, it also slightly distorts the
real value of the harmonic potentials \cite{HonDes10}.

In order to explain the proposed method, a few basic elements need
to be revisited. First, a \emph{goal} is a grid cell with the lowest
harmonic value ($\phi^{(k)}(x,\,y)=0$). This value is fixed and it
will never be affected by its neighboring values. An \emph{obstacle
}is any cluster of cells blocking a potential path towards the goal.
Its value is maximum ($\phi^{(k)}(x,\,y)=1$) and is also never affected
by its neighbors. However, its position may change -- e.g. in dynamic
environments. \emph{Free space }is any grid cell that does not contain
an obstacle or the goal. The value of the harmonic potential in the
free space is initialized with $0.5$ and is updated at each iteration.

A path to the goal is given by an index matrix, $M_{idx}(x,\,y)$, which
for every coordinate $( x,\,y )$ contains the index of the neighbor with
the lowest harmonic potential. That is,
\begin{equation}
    M_{idx}(x,y)=min[\phi^{(k)}(x+1,y),\phi^{(k+1)}(x-1,y),\phi^{(k)}(x,y+1),\phi^{(k+1)}(x,y-1)]\textrm{.} \label{eq:index-matrix}
\end{equation}

\subsection{Rubber Band Model}

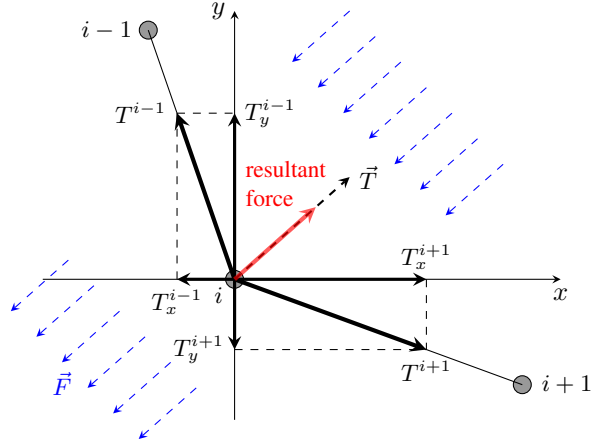
\begin{figure}
\centering{}
\begin{tikzpicture}[scale=0.85,>=stealth]
    \filldraw [fill=black!40,draw=black] (-1.35,3.9) circle[radius=0.14] node at (-2,3.9) {\small ${i-1}$};
    \filldraw [fill=black!40,draw=black] (4.5,-1.65) circle[radius=0.14] node at (5.2, -1.65) {\small ${i+1}$};
    \filldraw [fill=black!40,draw=black] (0,0) circle[radius=0.14] node at (-0.25,-0.25) {\small ${i}$};
    \draw [->] (-3,0) -- (5.1,0) node [right,below] {$x$};
    \draw [->] (0,-2.2) -- (0,4.2) node [above,left] {$y$};
    \draw [->, very thick] (0,0) -- (0,2.6) node [right] {\small $T_y^{i-1}$};
    \draw [->, very thick] (0,0) -- (0,-1.1) node [left] {\small $T_y^{i+1}$};
    \draw [->, very thick] (0,0) -- (-0.9,0) node [below] {\small $T_x^{i-1}$};
    \draw [->, very thick] (0,0) -- (3,0) node [above] {\small $T_x^{i+1}$};
    \draw [->, ultra thick] (0,0) -- (-0.9,2.6) node [left] {\small $T^{i-1}$};
    \draw [->, ultra thick] (0,0) -- (3,-1.1) node [below] {\small $T^{i+1}$};
    \draw [dashed] (3,0) -- (3,-1.1);
    \draw [dashed] (0,-1.1) -- (3,-1.1);
    \draw [dashed] (-0.9,0) -- (-0.9,2.6);
    \draw [dashed] (-0.9,2.6) -- (0,2.6);
    \draw (0,0) -- (-1.35,3.9);
    \draw (0,0) -- (4.5,-1.65);
    \draw [->, dashed, thick] (0,0) -- (1.8,1.6) node [right] {\small $\vec{T}$};
    \draw [->, ultra thick, red,opacity=0.65] (0,0) -- (1.27,1.13);
    \node at (0.2,1.5) [text width=1.2, align=center, red ]{\small resultant force};
    \draw [->, blue, dashed] (1.8,4.2) -- (0.9,3.4);
    \draw [->, blue, dashed] (2.2,3.8) -- (1.3,3.0);
    \draw [->, blue, dashed] (2.6,3.4) -- (1.7,2.6);
    \draw [->, blue, dashed] (3.0,3.0) -- (2.1,2.2);
    \draw [->, blue, dashed] (3.4,2.6) -- (2.5,1.8);
    \draw [->, blue, dashed] (3.8,2.2) -- (2.9,1.4);
    \draw [->, blue, dashed] (4.2,1.8) -- (3.3,1.0);
    \draw [->, blue, dashed] (-0.6,-1.7) -- (-1.5,-2.5);
    \draw [->, blue, dashed] (-1.0,-1.3) -- (-1.9,-2.1);
    \draw [->, blue, dashed] (-1.4,-0.9) -- (-2.3,-1.7);
    \draw [->, blue, dashed] (-1.8,-0.5) -- (-2.7,-1.3) node [below, blue] {\small $\vec{F}$};
    \draw [->, blue, dashed] (-2.2,-0.1) -- (-3.1,-0.9);
    \draw [->, blue, dashed] (-2.6,0.3) -- (-3.5,-0.5);
\end{tikzpicture}
\caption{{\small{}\label{fig:Tensions-exerted}Tensions exerted by consecutive cells along the path.}}
\end{figure}

As proposed in \cite{HonDes10}, the rubber band model is employed
to optimize the path obtained by the harmonic potential fields. This
idea of rubber band was previously introduced in \cite{Hilgert_03},
but mostly to define obstacle contours. Here, we combine the idea
of rubber band model and harmonic potentials to define the path as
a smoothed linked list of grid cells. The two immediately adjacent
cells in the link, i.e. the previous and the next cells along the
link from the current cell, exert internal forces on that same cell.
Figure \ref{fig:Tensions-exerted} illustrates this idea for the cell
$i$ and its previous and next cells in the path, $i-1$ and $i+1$,
respectively.

Every cell in the path is affected by two kinds of forces: the internal
tension (rubber band) forces $\vec{T}_{i\pm1}$, and
the potential force $\vec{F}$. The position of a cell
in the path is given by the pair $(x,\,y)$ that leads to the resultant
forces to be minimum. That is:
\begin{gather}
(\hat{x},\hat{y})=\underset{(x,\,y)}{argmin}\,\left\Vert \vec{F}+\vec{T}_{i+1}+\vec{T}_{i-1}\right\Vert \label{eq:resultant} \\
(x,\,y)_{k+1}=(x,\,y)_{k}+\delta{(\hat{x},\hat{y})} \label{eq:drag}
\end{gather}
where $\delta{(\hat{x},\hat{y})}$ represents the direction toward the pair $(\hat{x},\hat{y})$ in Equation (\ref{eq:resultant}), used to move
the position of the current cell in the path at each iteration.

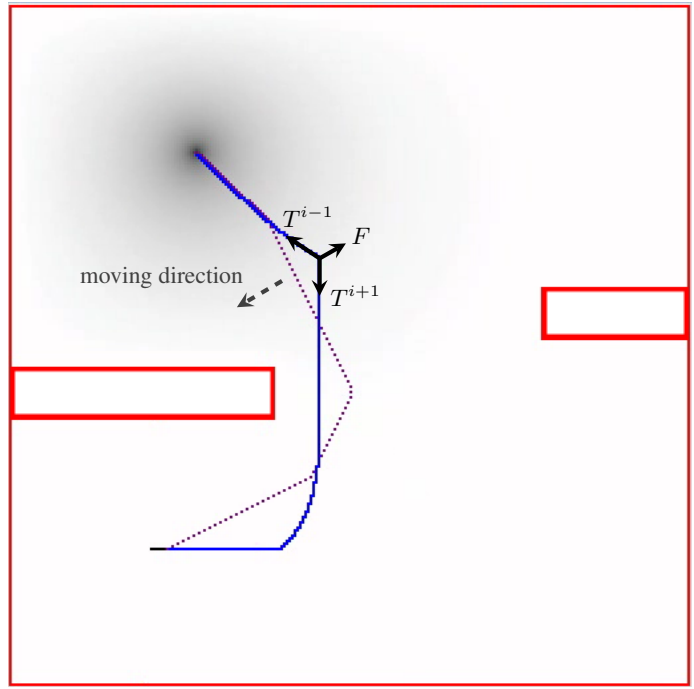
\begin{figure}
    \centering
    \begin{tikzpicture}[>=stealth]
    \node[inner sep=0pt] (harmonic) at (0,0)
    {\includegraphics[width=.55\textwidth,trim={5 5 5 25},clip]{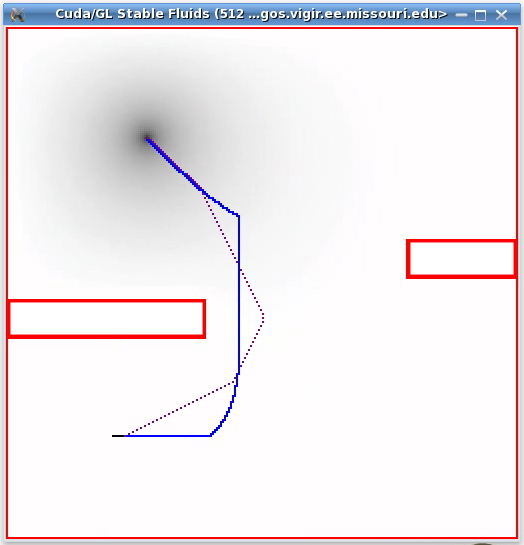}};
    \draw [->, ultra thick] (-0.4,1.15) -- (-0.4,0.65) node [right] {\footnotesize $T^{i+1}$};
    \draw [->, ultra thick] (-0.4,1.15) -- (-0.85,1.45) node at (-0.55,1.7) {\footnotesize $T^{i-1}$};
    \draw [->, ultra thick] (-0.4,1.15) -- (-0.05,1.35) node at (0.15,1.45) {\footnotesize $F$};
    \draw [->, ultra thick, dashed, darkgray] (-0.9,0.85) -- (-1.5,0.5) node at (-2.5,0.9) {\footnotesize moving direction};
    \end{tikzpicture}
    \caption{{\small Here, it is shown the resultant
forces on cell $i$, as well as the harmonic potential path (full/blue
line) being optimized by the rubber band model (dotted/purple line).}}
    \label{fig:Resultant-forces}
\end{figure}

Let us assume that the current coordinates of the $i^{th}$
cell is $(x_{i},\,y_{i})$, and the coordinates of the two neighbors
are $(x_{i-1},\,y_{i-1})$ and $(x_{i+1},\,$ $y_{i+1})$, respectively. The
resultant of the forces on the cell $i$, as shown by Figure \ref{fig:Resultant-forces},
provides the direction and intensity with which the path should be
moved in order for the forces to reach equilibrium. 

The last component of these calculations is the force $\vec{F}$
derived from the harmonic potential. This force is calculated using:
\begin{equation}
\vec{F}=(1-\phi^{(k)}(\hat{x},\,\hat{y}))^{-1}-(1-\phi^{(k)}(x_{i},\,y_{i}))^{-1} \label{eq:force}
\end{equation}
where $\phi^{(k)}(x_{i},\,y_{i})$ represents the harmonic potential at the current position of the cell in the path given by Equation (\ref{eq:discretelaplace}), and $\phi^{(k)}(\hat{x},\,\hat{y})$ represents the harmonic potential of the candidate position to which the cell $i$ is being dragged.
At each step, the robot moves towards the next optimized position
in the path.

Figure \ref{fig:Resultant-forces} summarizes the idea of the harmonic
potentials and the internal tension (elastic) forces of the model.
In the figure, red lines represent obstacles (walls) and the black
area is the desired destination of the robot. The darker the color
in free space, the lower the harmonic potential value. Figure \ref{fig:Resultant-forces}
also shows how the path obtained from the simple application of harmonic
fields (full/blue path) compares to the one being optimized by the
rubber band model (dotted/purple path).

\subsection{Kalman Filter}

A Kalman filter addresses the general problem of estimating the state
$x\:\in\mathbb{R}^{n}$ of a discrete-time process, which in our case represents
the position of a moving obstacle. This state evolves with the step
$k$, governed by the following linear stochastic equation \cite{WelBis97}:
\begin{equation}
x_{k}=A_{k}x_{k-1}+B_ku_{k-1}+w_{k-1}\label{eq:kalman_state}
\end{equation}
where $A$, the $n\times n$ state matrix, relates the state at time
step $k-1$ to the state at $k$; $B$, the $n\times l$ input matrix,
relates the control input $u\in \mathbb{R}^{l}$ to the state $x$, and the
random variable $w_{k}$ represents a process noise. Due to sensor
limitations, the state of the system is observed by $z\:\in\mathbb{R}^{m}\;\;(m<n)$
according to a measurement model given by: 
\begin{equation}
z_{k}=H_{k}x_{k}+v_{k}\label{eq:kalman_measurement}
\end{equation}
where the $m\times n$ matrix H is the observation matrix, and the random variable $v_{k}$ represents
the uncertainty of the observation given the current state. Also,
all random variables in the Kalman filter are assumed to be independent, normal
distributions. That is, $p(x_{i}|z_{j})\sim N(\mu_{i|j},P_{i|j})$
for $i,j\in\{k-1,k\}$, $p(z_{i}|x_{j})\sim N(\lambda_{i|j},S_{i|j})$
for $i,j\in\{k-1,k\}$, $p(w)\sim N(0,Q)$, and $p(v)\sim N(0,R)$.

The Kalman filter is an iterative method that alternates between
two phases: prediction and update. Since in our case, moving obstacles
are assumed to move without requiring any input $u$, at each iteration,
the Kalman filter predicts the system's next state through Equations (\ref{eq:kalman_time1}) and (\ref{eq:kalman_time2}):
\begin{equation}
\hat{x}_{k|k-1}=A_{k}\hat{x}_{k-1|k-1}\label{eq:kalman_time1}
\end{equation}
\begin{equation}
P_{k|k-1}=A_{k}P_{k-1|k-1}A_{k}^{T}+Q_{k-1}\label{eq:kalman_time2}
\end{equation}
where $\hat{x}_{k|k-1}$ is the estimation of the state $x$ based on the observations up to time $k-1$, and the estimate variance $P_{k|k-1}$ is the mean squared error in the estimate $\hat{x}_{k|k-1}$.

Then during the update phase, the state of the system is refined by:
\begin{equation}
\hat{x}_{k|k}=\hat{x}_{k|k-1}+K_{k}(z_{k}-H_{k}\hat{x}_{k|k-1})\label{eq:kalman_measurement1}
\end{equation}
\begin{equation}
P_{k|k}=(I-K_{k}H_{k})P_{k|k-1}\label{eq:kalman_measurement2}
\end{equation}
where the Kalman gain, $K_{k}$ is defined as 
\begin{equation}
K_{k}=P_{k|k-1}H_{k}^{T}(H_{k}P_{k|k-1}H_{k}^{T}+R_{k})^{-1} \textrm{ .}\label{eq:kalman_measurement3}
\end{equation}

The integration of Kalman filter and the time-warped grid will be discussed in Section
\ref{subsec:Integrating-KF-HP-TWG}.

\subsection{Time-Warped Grid\label{subsec:Time-Warps}}

In a static environment, a path computed by combining the harmonic
potentials and the rubber band model above is guaranteed to always
exist and to be smooth. However, if obstacles are allowed to freely
move in the environment, that same path must be constantly updated
using the robot sensors (in our case the laser range sensor and the vision system). In that
case, newly sensed positions of the obstacles will lead to new harmonic
potentials and hence new smoothened paths to the goal from the rubber
band model. This behavior is undesired, since it can force the robot
to follow inefficient paths like loops, paths that sharply move the
robot away from the goal, and/or paths that bring the robot dangerously
close to obstacles. By using time-warped grid and Kalman filter to
estimate obstacle velocities, our system is able to estimate future
positions of the obstacles and calculate optimum paths despite of
moving obstacles.

The idea of time-warped grid was inspired in general relativity,
where the fabric of space-time is warped by large gravitational forces.
Here, instead of gravitational forces, we use the velocities of the
robot and of the obstacles to warp the grid in the environment map.
The motivation for that, as we will explain in greater detail next,
comes from the consequent assignment of a parallel processor for the
computation of the potentials at each cell of the grid -- and hence
the path of the robot to the goal. Since the grid is warped by the
velocities of all moving obstacles in the grid, the path itself becomes
a function of those velocities, leading to a path plotted for the
future position of those obstacles, rather than the current ones.

\begin{figure}[H]
\centering{}
\begin{tikzpicture}[>=stealth]
    \node[inner sep=0pt] (grid) at (0,0)
    {\includegraphics[width=0.5\columnwidth,trim={115 45 95 30},clip] 
    {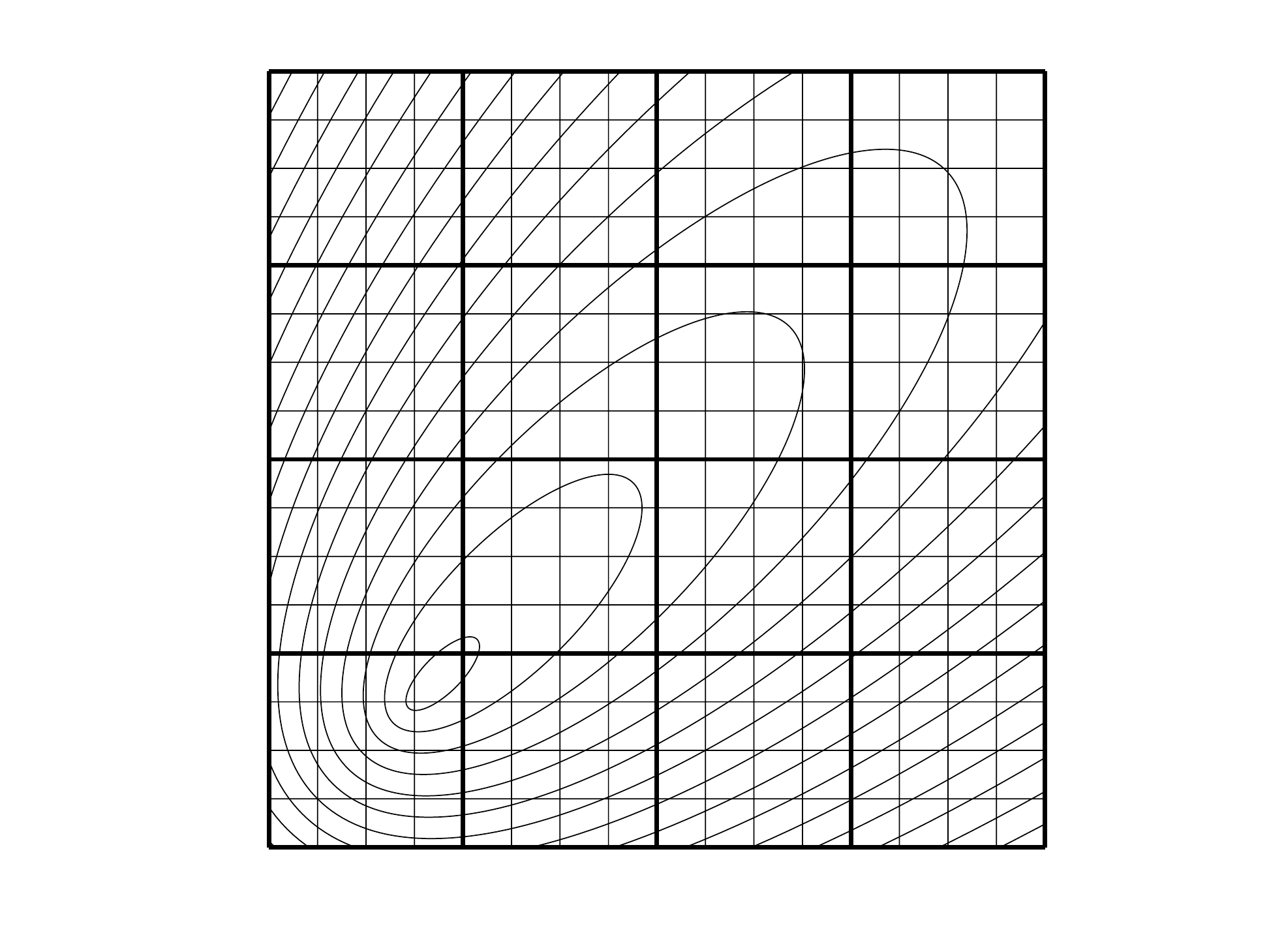}};
    \node[inner sep=0pt, rotate=45] (robot1) at (-2.3,-2.3)
    {\includegraphics[scale=0.6,trim={1 1 1 1},clip] {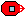}};
    \node[inner sep=0pt, rotate=-140] (robot2) at (1.7,1.7)
    {\includegraphics[scale=0.6,trim={1 1 1 1},clip] {robot_screenshot.png}};
    \node[inner sep=0pt, rotate=-15] (robot3) at (-2.9,1.5)
    {\includegraphics[scale=0.6,trim={1 1 1 1},clip] {robot_screenshot.png}};
    \draw [<-, thick, blue] (-1.8,-2.3)--(-0.1,-2.3) node [right] {\small \textbf{main robot}};
    \draw [<-, thick, blue] (-2.6,1.8)--(-1.0,3.0) node [above] {\small \textbf{moving obstacles}};
    \draw [<-, thick, blue] (1.45,1.85)--(-0.5,3.0);
\end{tikzpicture}
\caption{{\small{}\label{fig:warps} Time-Warped Grid created by concentric
ellipses around the main robot.}}
\end{figure}

First, imagine a grid warped by the velocity
of the robot alone. Such grid, depicted in Figure \ref{fig:warps},
has enlarged squares in front of the robot, i.e. in the direction
of motion. If we think of these squares as the space traveled in one
unit of time, it becomes obvious that the grids must be larger in
front of the robot, smaller behind it, and the same on each side.
In fact, in the direction of the motion of the robot, the warped grids
stretch like ellipses centered around the position of the robot. Each
of such ellipse can be numbered, representing the degree of warping.
In order to estimate the future positions of moving
obstacles, the robot should find the warps containing moving obstacles
and label the obstacle with the corresponding warp number. As a result,
if an obstacle moves towards the robot, its label decreases. 

The time-warp of a moving obstacle with position of $( x_{obj},\,y_{obj} )$ is derived using the equation for a standard ellipse which has been rotated through an angle $\theta$:
\begin{align}
    & \frac{\big(\cos{(\theta)}(x_{obj}-x_c)+\sin{(\theta)}(y_{obj}-y_c)\big)^2}{r_x^2}+ \nonumber \\
    & \frac{\big(\sin{(\theta)}(x_{obj}-x_c)-\cos{(\theta)}(y_{obj}-y_c)\big)^2}{r_y^2} = 1 \label{eq:ellipse}
\end{align}
where $(r_x,\,r_y)$ are the major and minor axes of the ellipse respectively, and $\theta$ is the orientation of the main robot.   The centers of ellipses $(x_c,\,y_c)$ are calculated based on the position of the main robot $(x_r,\,y_r)$, so the ellipses are larger and longer as they get farther from the robot, that is:
\begin{equation*}
    x_c=x_r+0.9\,r_x\cos{\theta} \quad \textrm{and} \quad y_c=y_r+0.9\,r_x\sin{\theta}\textrm{.}
\end{equation*}

The elliptical grids are set for this application such that $r_x=4r_y$. Therefore, the time-warp number of a moving obstacle is calculated as the length of the major axis of the ellipse on which the moving obstacle is located:
\begin{align}
    r_x= & \Big(\big(\cos{(\theta)}(x_{obj}-x_c)+\sin{(\theta)}(y_{obj}-y_c)\big)^2 \nonumber \\ & +16\big(\sin{(\theta)}(x_{obj}-x_c)-\cos{(\theta)}(y_{obj}-y_c)\big)^2\Big)^{\frac{1}{2}} \label{eq:time-warp}
\end{align}
After calculating the time-warps, Kalman filter
estimates the future position of the moving obstacles for the next $j$'th
step, where $j$ is the warp number of the grid occupied by the detected
obstacle. Needless to say, Kalman filter algorithm should be initialized
with respect to the velocity of the robot and the amount of noise in
sensor readings.

The main idea of time-warped grid is to acknowledge that the further
the obstacle is, the more delayed should be its impact in the path
unless the obstacle also moves with great velocity towards the path.
In summary:
\begin{equation}
\textrm{Time-Warp}\,=:\,v \cdot t
\end{equation}
where $v$ is the ratio of the velocity of the robot to the moving
obstacle and $t$ is the warp number i.e. the corresponding ellipse
number.
\begin{equation}
( x,\,y )_{\textrm{future}}=( x,\,y )_{\textrm{estimated by Kalman filter based on time-warp}}
\end{equation}

This estimated position of the obstacle is considered as the possible
collision point with the robot and marked on the grid as if it were
a fixed obstacle by increasing the corresponding harmonic value. Therefore,
this obstacle would only affect the path of the robot if its future
position lies in the vicinity of the robot path. Since the further
the distance of the robot to the warped grid, the worst is Kalman
filter predictions -- i.e. the further in the future is the Kalman
filter prediction, the less accurate it is -- the assignment of ``occupied''
grids (high harmonic values) uses a Gaussian function. In other words,
not only the estimated positions of the obstacles in the future are
marked as occupied, but also their neighboring points. This Gaussian
function is defined based on the calculated uncertainty given by Kalman
filter at each step (Equation (\ref{eq:kalman_measurement2})) and a desired
``safety'' distance to obstacles.

At each iteration of the algorithm, the grid is cleared from harmonic
potentials, and new values are updated onto the grid. This approach
is justified since the estimations from the Kalman filter do not change
drastically from one iteration to the next, and the values evolve
slowly anyways. Also, at each iteration, the predicted harmonic potentials
are used together with the rubber band model to determine a path for
the robot to follow. It should be mentioned here that the robot does not have any prior
knowledge about the environment, except for the position of the goal.

It should be clear to the reader by now, that the time-warped grid
method reduces and simplifies many calculations. For one, it eliminates
the need to take the directions of the movement and the absolute value
of the distance between the robot and the moving obstacles into consideration
for the calculation of the path. In fact, the degree of warping assigned
to the grids encodes both that direction and the time factors required
for the calculations of the path.

In the next section, it will be further detailed the use of the time-warped grid combined with the Kalman filtering. 

\subsection{Integration of Kalman Filter with Time-Warps\label{subsec:Integrating-KF-HP-TWG}}

As mentioned before, in this work the estimation of the position of
moving obstacles in the future is done by Kalman filter. It is assumed
that the current position of the obstacle is not known accurately
due to noise and other errors in the vision-based localization system. In fact, the motion
dynamics of the obstacles and their associated amount of noise are modeled in the implemented Kalman filter. The estimated positions
of the obstacles are based on the warp number, the dynamic model and
the previous positions of the obstacle using the iterative process
described by Equations (\ref{eq:kalman_state} - \ref{eq:kalman_measurement3}).

For the experiments presented in Section \ref{sec:Experimental-Results}, the moving obstacles are simulated by the MobileSim software and the Aria API \cite{Whitbrook09},
and they wander inside of a two-dimensional map with a mostly constant
velocity\footnote{Interactions between moving obstacles can lead to change of velocity
and/or direction of motion by one or all obstacles involved.}. Therefore, the state $x$ contains $x_r$ and $y_r$ positions, and
$\dot{x}_r$ and $\dot{y}_r$ velocities, and $A$ matrix in Equation (\ref{eq:kalman_state}) is formed as below:
\begin{equation*}
x=\left[\begin{array}{c}
x_r\\
y_r\\
\dot{x}_r\\
\dot{y}_r
\end{array}\right] , \; A=\left[\begin{array}{cccc}
1 & 0 & dt & 0\\
0 & 1 & 0 & dt\\
0 & 0 & 1 & 0\\
0 & 0 & 0 & 1
\end{array}\right]
\end{equation*}

There is no input for this application so $B$ matrix is zero in Equation
(\ref{eq:kalman_state}). The vision-based localization system provides measurements
only from $x_r$ and $y_r$ position, not the velocities. Therefore, the $H$ matrix in Equation (\ref{eq:kalman_measurement}) becomes:
\begin{equation*}
H=\left[\begin{array}{cccc}
1 & 0 & 0 & 0\\
0 & 1 & 0 & 0
\end{array}\right] \textrm{ .}
\end{equation*}

Since the simulated moving obstacles usually have only linear motion
(they turn only when they get close to walls and other obstacles),
the process noise $w_{k}$ ($Q$) is assumed to be very small. On
the other hand, observations can be noisy and since we have no knowledge
about the exact position of the obstacles, the measurement noise $v_{k}$
($R$) should be adequately set.

As we mentioned before, at each step, the Kalman filter uses the computed a-posteriori
error covariance estimate ($P_{k}$) to adjust the uncertainty for
the neighbors of the estimated position. In other words, the more
uncertainty returned by the Kalman filter, the larger is the marked
area on the grid around the estimated position of the obstacle.

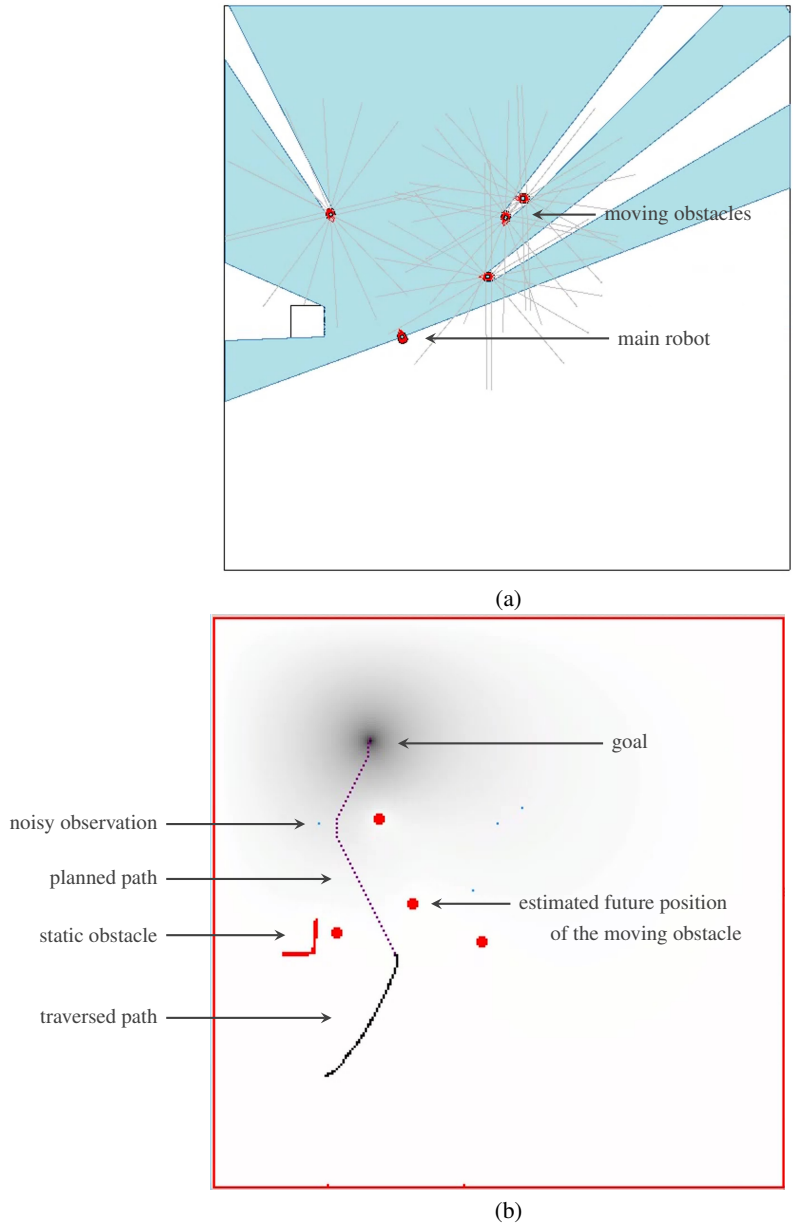
\begin{figure}
\begin{tabular}{c}
\begin{tikzpicture}[>=stealth, scale=0.89, every node/.style={scale=0.89}]
    \node[inner sep=0pt] (harmonic) at (0,0)
    {\includegraphics[width=0.525\columnwidth,trim={5 5 5 5},clip]{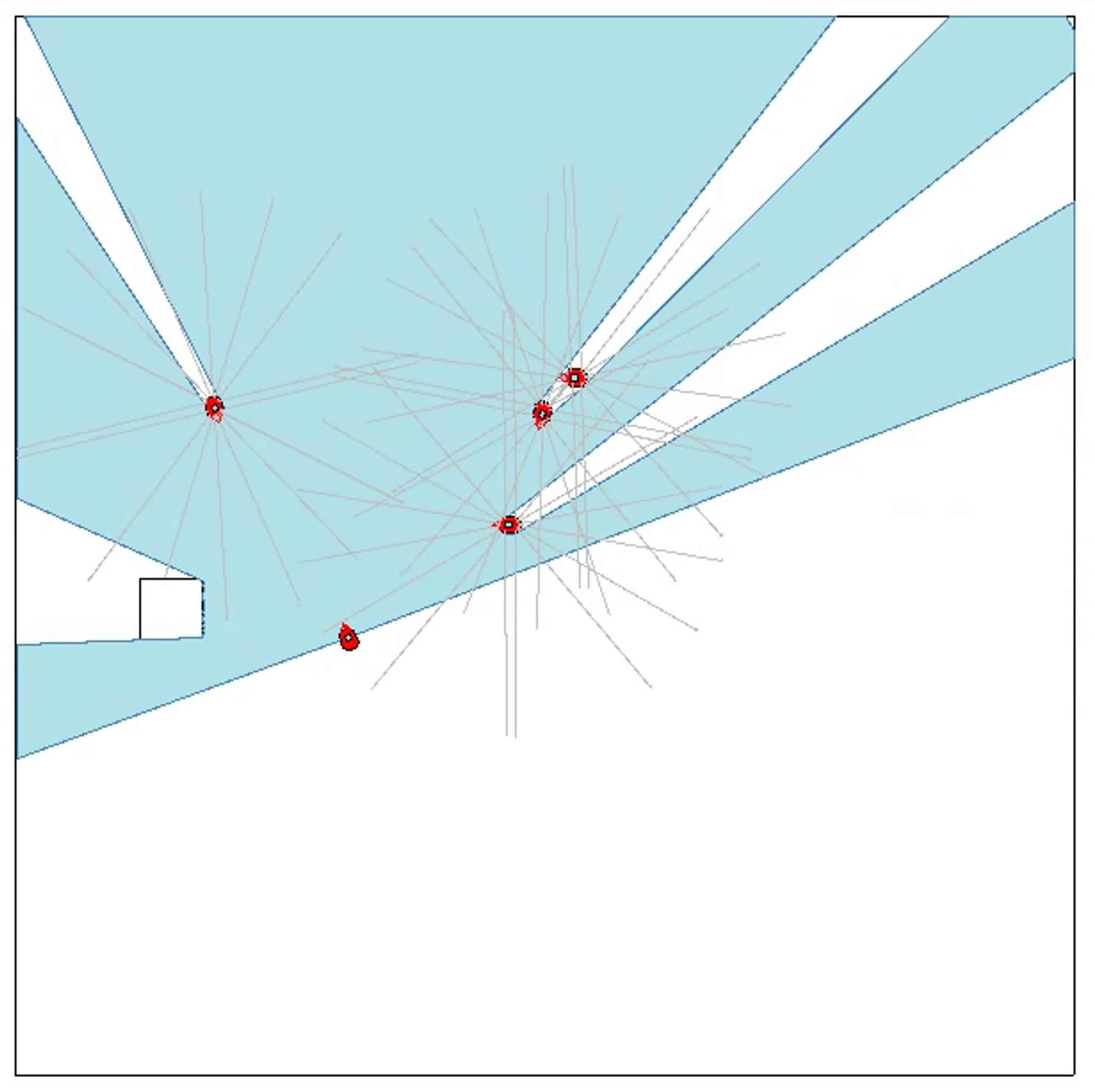}};
    \draw [<-, thick, darkgray] (-1.25,-0.75)--(1.5,-0.75) node [right] {\small main robot};
    \draw [<-, thick, darkgray] (0.35,1.1)--(1.3,1.1) node [right] {\small moving obstacles};
\end{tikzpicture}\tabularnewline
{\footnotesize (a)}\tabularnewline
\hspace*{-9em}
\begin{tikzpicture}[>=stealth, scale=0.88, every node/.style={scale=0.88}]
    \node[inner sep=0pt] (harmonic) at (0,0)
    {\includegraphics[width=0.525\columnwidth,trim={3 3 3 65},clip]{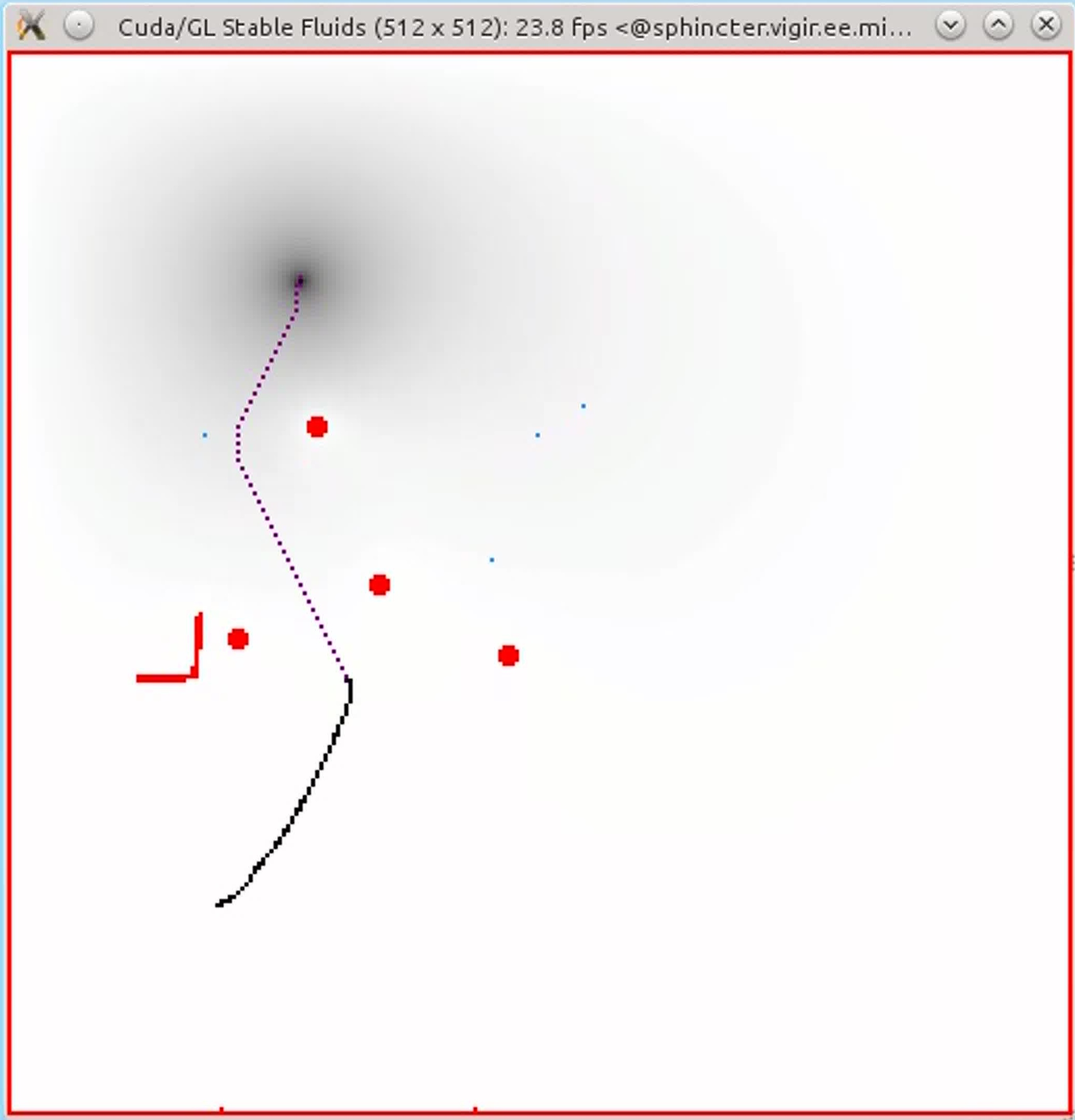}};
    \draw [<-, thick, darkgray] (-1.5,2.4)--(1.6,2.4) node [right] {\small goal};
    \draw [<-, thick, darkgray] (-2.9,1.18)--(-5,1.18) node [left] {\small noisy observation};
    \draw [<-, thick, darkgray] (-2.5,0.35)--(-5,0.35) node [left] {\small planned path};
    \draw [<-, thick, darkgray] (-1,-0.025)--(0.2,-0.025) node [right] {\small estimated future position};
    \node [darkgray] at (2.25,-0.5) {\small of the moving obstacle};
    \draw [<-, thick, darkgray] (-3.2,-0.5)--(-5,-0.5) node [left] {\small static obstacle};
    \draw [<-, thick, darkgray] (-2.5,-1.75)--(-5,-1.75) node [left] {\small traversed path};
\end{tikzpicture}\tabularnewline
{\footnotesize (b)}\tabularnewline
\end{tabular}
\centering{}\caption{{\small{}\label{fig:good-explain}The effect of the moving obstacles
on the planned path: (a) actual environment (b) program's internal representation, which shows the estimated and noisy observations made by the laser sensor and the vision-based localization system (small blue dots), the predicted future positions of the moving obstacles by time-warps and Kalman filter (larger red dots), the detected static obstacles (red lines), the planned path (purple dashed line), the traversed path (black solid line), and the goal position (darkest area indicating the lowest harmonic potential value).}}
\end{figure}

Figure \ref{fig:good-explain} illustrates how the estimated future positions based on the time-warped grid and Kalman filtering affect
the planned path by harmonic potential fields and the rubber band
model, in order to avoid collisions.
The window in Figure \ref{fig:good-explain}(a) is the output of MobileSim with the actual robots, the laser range sensor information, and the walls of the environment. The window in Figure \ref{fig:good-explain}(b)
is created by the main program using OpenGL to present the internal representation of the environment based on the sensed information. As the reader can notice, Figure \ref{fig:good-explain}(b) presents some small blue dots, besides the larger red dots.  The first are the estimated and noisy observations made by the laser sensor and the vision-based localization system, while the latter are the predicted future positions of the moving obstacles by time-warps and Kalman filter. The detected static obstacles are marked with red lines. The solid black line represents the traversed path by the robot, while the dotted purple line shows the planned path by harmonic potentials and rubber band optimization. The goal position is represented by the darkest area on the map, indicating the lowest harmonic potential value.

\subsection{GPU-Based Parallel Implementation}

All the above mentioned algorithms are highly computationally intensive especially when the map is large and/or there are many moving obstacles in the environment. The problem of using harmonic fields is that it requires repeated updates of the potential values at every cell of the grid. These updates are in turn a function of the potential of the neighboring cells, which leads to a recursive and quite-time consuming
algorithm. Furthermore, the algorithm for optimizing
the path using the rubber band model needs to calculate different
forces at every cells of the grid in all iterations which is a heavy
process. Estimating the future positions of moving obstacles  requires another real-time process using time-warps and
Kalman filter for each moving obstacle. An effective way to speed
up these time consuming algorithms is through the use of parallel
computation. Even more pertinent to our problem, since a fine and
detailed grid of the environment may require millions of cells, we
propose the parallelization through the use of General Purpose Graphics
Processor Units (GPGPUs). Due to the use of grid-based maps, all these
algorithms translate quite nicely into the parallel paradigm of GPU
computing. 

In that sense, several CUDA programs, namely kernel functions, are implemented to carry out the calculation of grid cells. The environment is divided into 10 cm by 10 cm cells,
and each cell contains the input data for a GPU kernel called by
a number of threads equal to the number of grid cells. In other words,
each GPU thread is responsible for its own cell of the grid -- e.g.
for a 25.6 meter by 25.6 meter area, 256 $\times$ 256 threads (forming
16 by 16 thread blocks) are created to process the data in the grids. 
To this end, four GPU kernels are called in order to: 1) calculate the warp of each cell; 2) if there is
a moving obstacle in the cell, run the Kalman filter algorithm to
estimate the future position of the obstacle; 3) calculate the harmonic
potential value of each cell; and 4) optimize the path using rubber
band model. Each kernel function is multiply instantiated by the CUDA platform and the calculations of the various grid cells is performed in parallel by the GPU. 

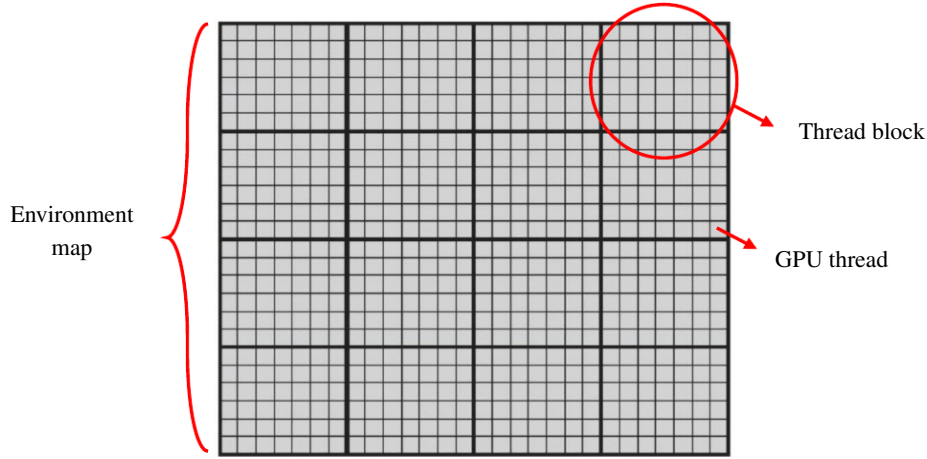
\begin{figure}
\centering{}
\begin{tikzpicture}[>=stealth]
\node[inner sep=0pt] (harmonic) at (0,0)
{\includegraphics[width=0.5\columnwidth,trim={150 10 115 0},clip]{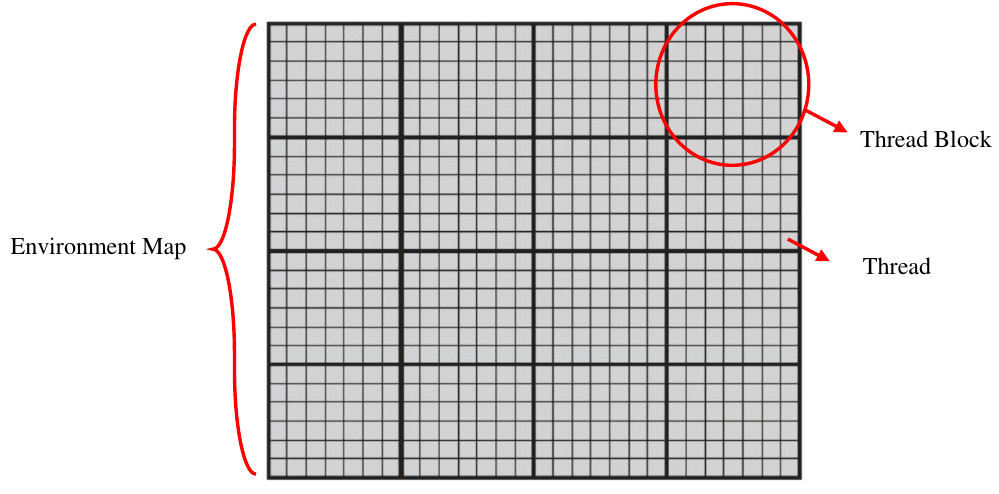}};
\node at (-5.2,0.2) {\small Environment};
\node at (-5.2,-0.3) {\small map};
\node at (5.3,1.3) {\small Thread block};
\node at (4.9,-0.4) {\small GPU thread};
\end{tikzpicture}
\caption{{\small{}\label{fig:map_squares}Mapping the environment to a GPU. Each grid cell is assigned to a GPU thread. 
The thick lines show the boarder of the thread blocks.}}
\end{figure}

Using different thread blocks provides the advantage of using GPU
shared memory. The calculated harmonic values reside in the GPU global
memory and are accessed by different threads. Since the neighbor cells
have some shared harmonic values, by loading the harmonic values into
the shared memory, the number of global memory accesses are reduced,
which has a significant effect on the speed of the algorithm. An example
of an environment divided into grid cells and their corresponding
threads and thread blocks are shown in Figure \ref{fig:map_squares}.
The figure illustrates this advantage of thread blocks and their shared
memory.

Moreover, using GPUs as proposed enables the robot to handle many (if
not all) moving obstacles at the same time. Some test scenarios with
large number of obstacles can be found in Section \ref{sec:Experimental-Results}. Final control of the robot and wandering the moving obstacles inside of the map are done by CPU.

The proposed path planning procedure is outlined in Algorithm \ref{alg:algorithm}.

\begin{algorithm}
\caption{Dynamic Path Planning via Time-Warped Grids}\label{alg:algorithm}
\begin{algorithmic}[1]
\footnotesize
\Require{Initialize the 2D map and harmonic potential values}
\Procedure{Map Update}{}
\State Read laser range sensor and vision-based localization data
\State Detect static and dynamic obstacles
\For {each object} \textbf{in parallel}
\If{Static}
    \State Update the harmonic potential value: $\phi(x,y)=1$
\ElsIf{Dynamic}
    \State Calculate the time-warp number of the moving obstacle via Eq. \ref{eq:time-warp}
    \State Estimate the future position $(x,y)_{future}$ using Kalman filter via Eqs. \ref{eq:kalman_state}-\ref{eq:kalman_measurement3} 
    \State Update the the harmonic potential value: {\small $\phi((x,y)_{future})=1$}
\EndIf
\EndFor
\EndProcedure
\Procedure{Path Planning}{}
\For {each cell in the map} \textbf{in parallel}
\For {k = 1 to 100}
\State Calculate harmonic potentials $\phi^{(k)}(x,y)$ via Eq. \ref{eq:harmonic-potential} 
\EndFor
\EndFor
\For {each cell in the map} \textbf{in parallel}
\State Update the index matrix with the position of the neighbor with lowest harmonic potential value via Eq. \ref{eq:index-matrix} 
\EndFor
\For {each cell in the index matrix} \textbf{in parallel}
\State Calculate tension force $\vec{T}$ and potential force $\vec{F}$ via Eq. \ref{eq:force}
\State Update the index matrix with the position of the neighbor with the minimum resultant force (rubber band optimization via Eqs. \ref{eq:resultant} and \ref{eq:drag}) 
\EndFor
\State Generate the current path based on the optimized index matrix starting from the current position of the robot
\State Move the robot to the next cell on the current path
\EndProcedure
\end{algorithmic}
\end{algorithm}

\section{Experimental Results \label{sec:Experimental-Results}}

Various test scenarios were performed using different maps and conditions
of the environments, as well as the number of moving obstacles. All
tests were performed for indoor navigation using a robot simulator
for the Pioneer P3-DX robots (MobileSim). It is important to mention
that all test scenarios were conducted using the NVIDIA GeForce GTX480
GPU and the CUDA programming environment.

\begin{figure}
\centering
\begin{tabular}{cc}
\includegraphics[width=5.2cm]{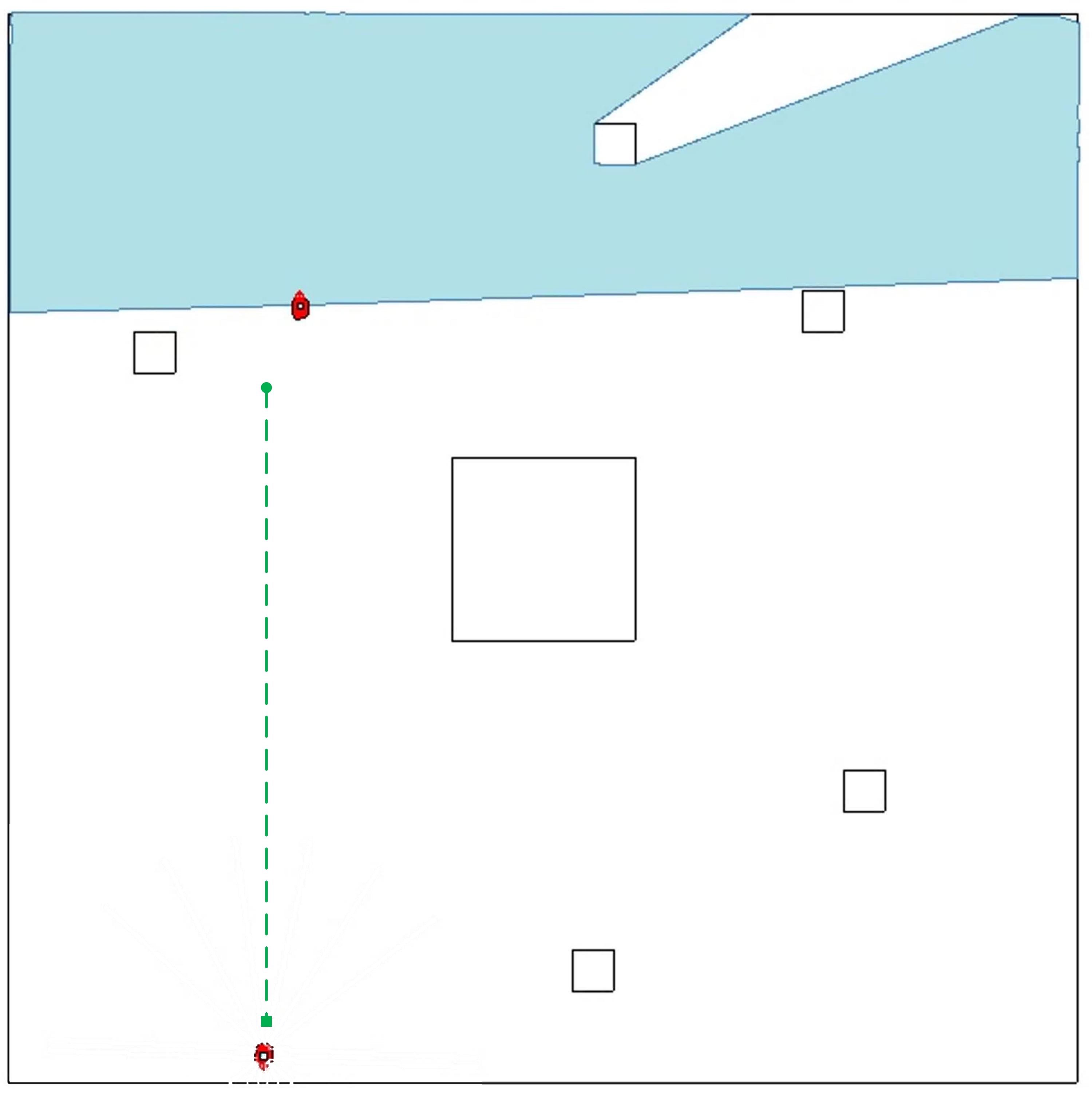} & \includegraphics[width=5.2cm,trim={10 4 10 90},clip]{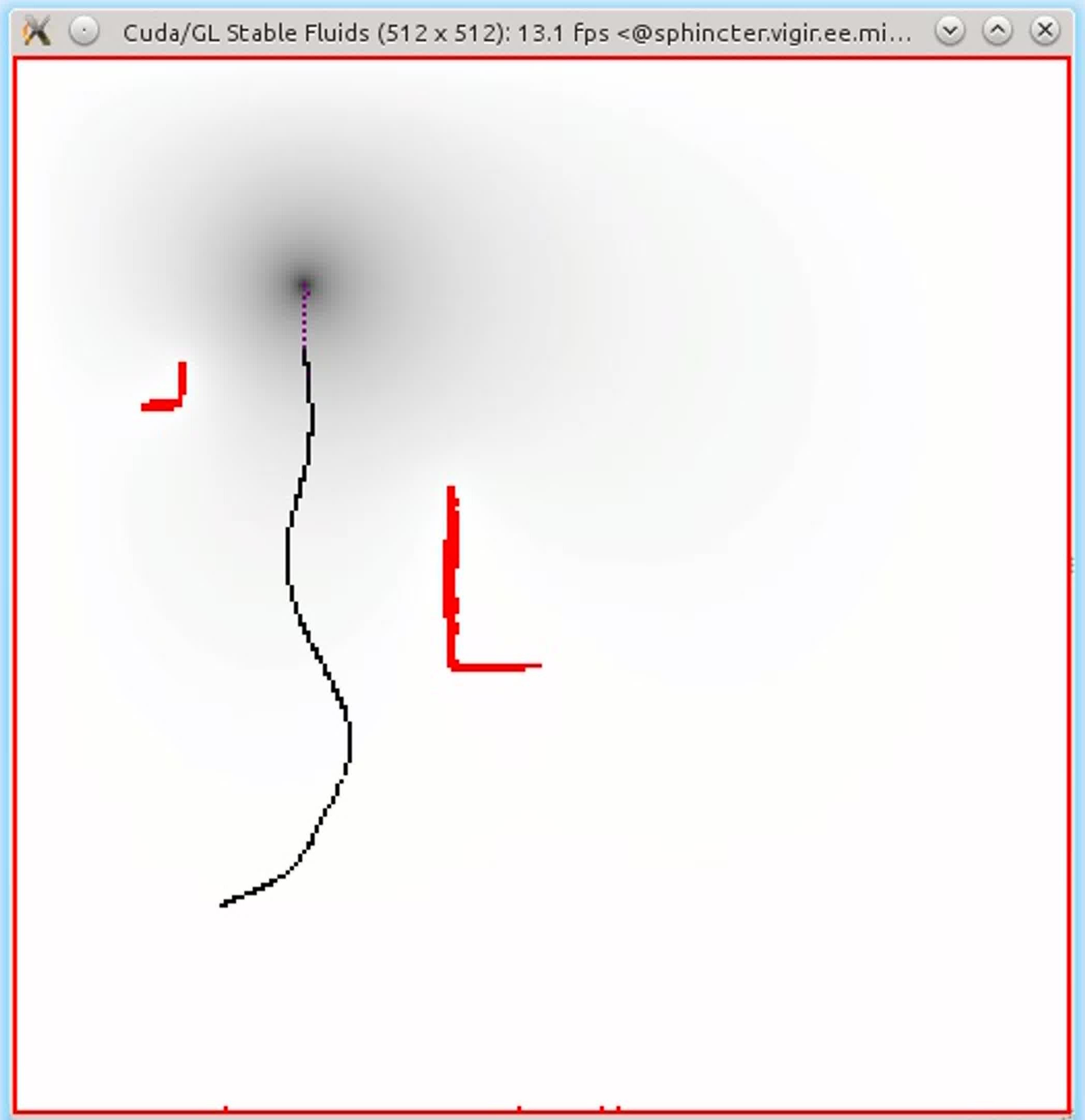}\tabularnewline
\multicolumn{2}{c}{{\small{}(a) 1 moving obstacle, map $\# $ 1.}}\tabularnewline
\includegraphics[width=5.2cm]{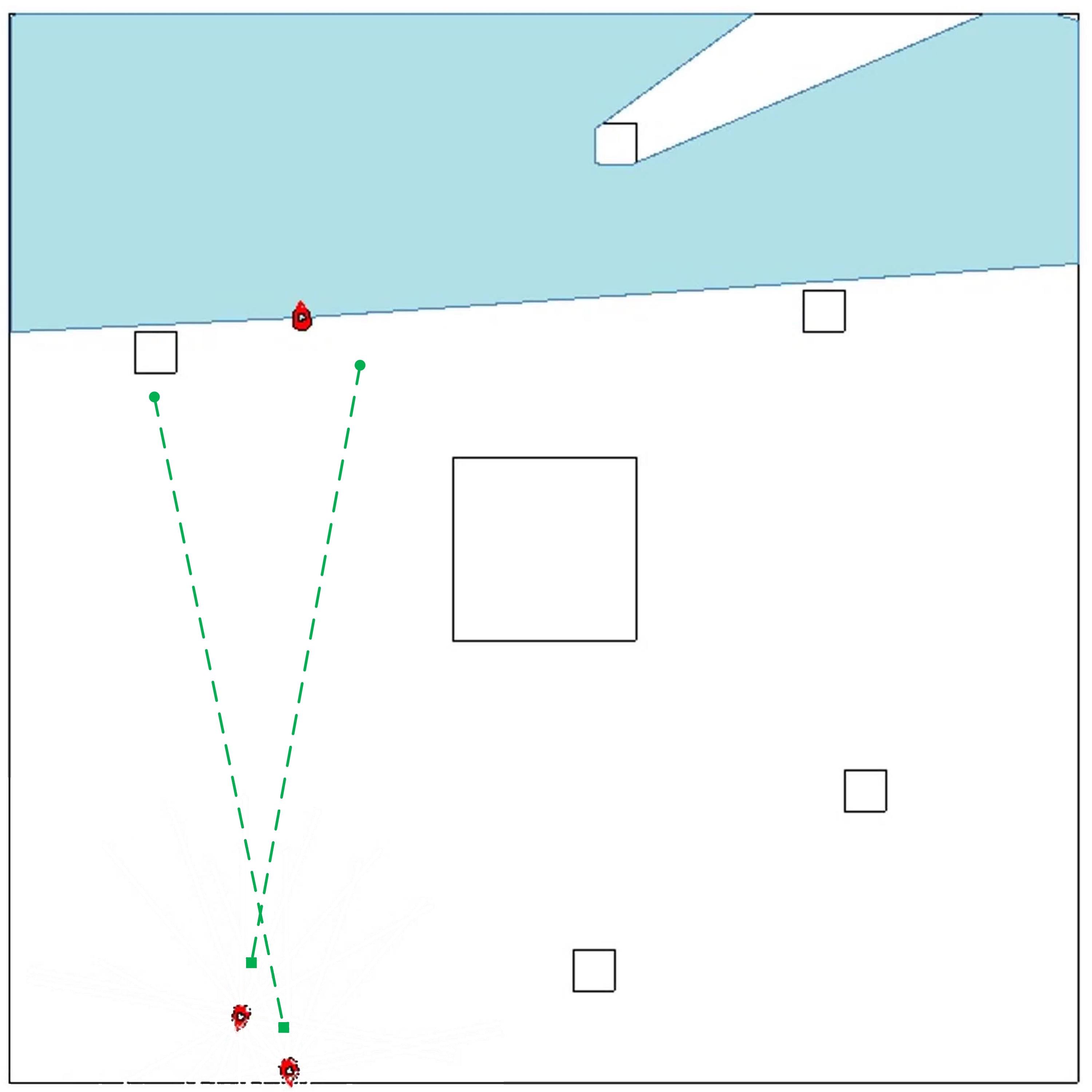} & \includegraphics[width=5.2cm,trim={10 4 10 90},clip]{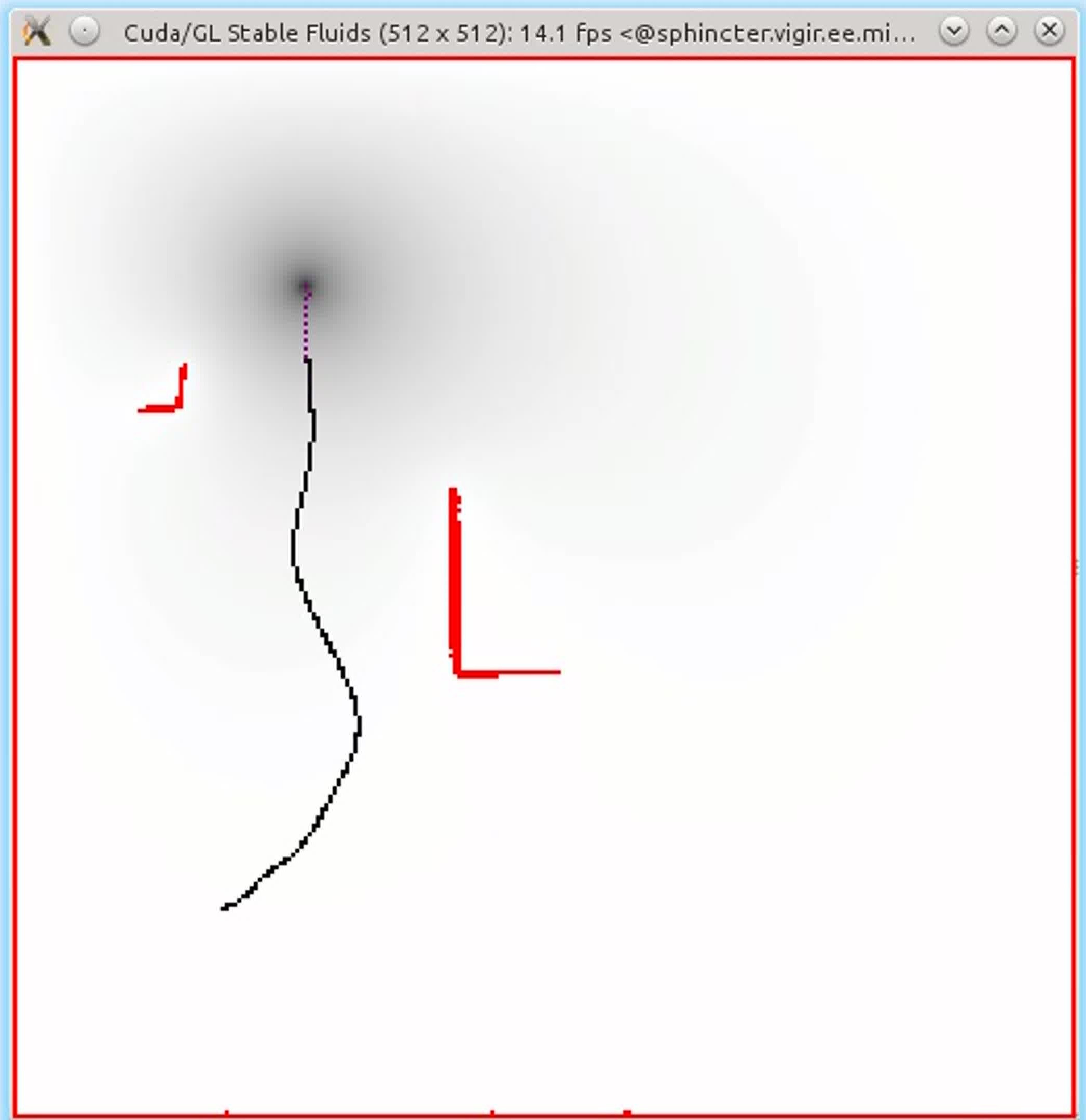}\tabularnewline
\multicolumn{2}{c}{{\small{}(b) 2 moving obstacles, map $\# $ 1.}}\tabularnewline
\includegraphics[width=5.2cm]{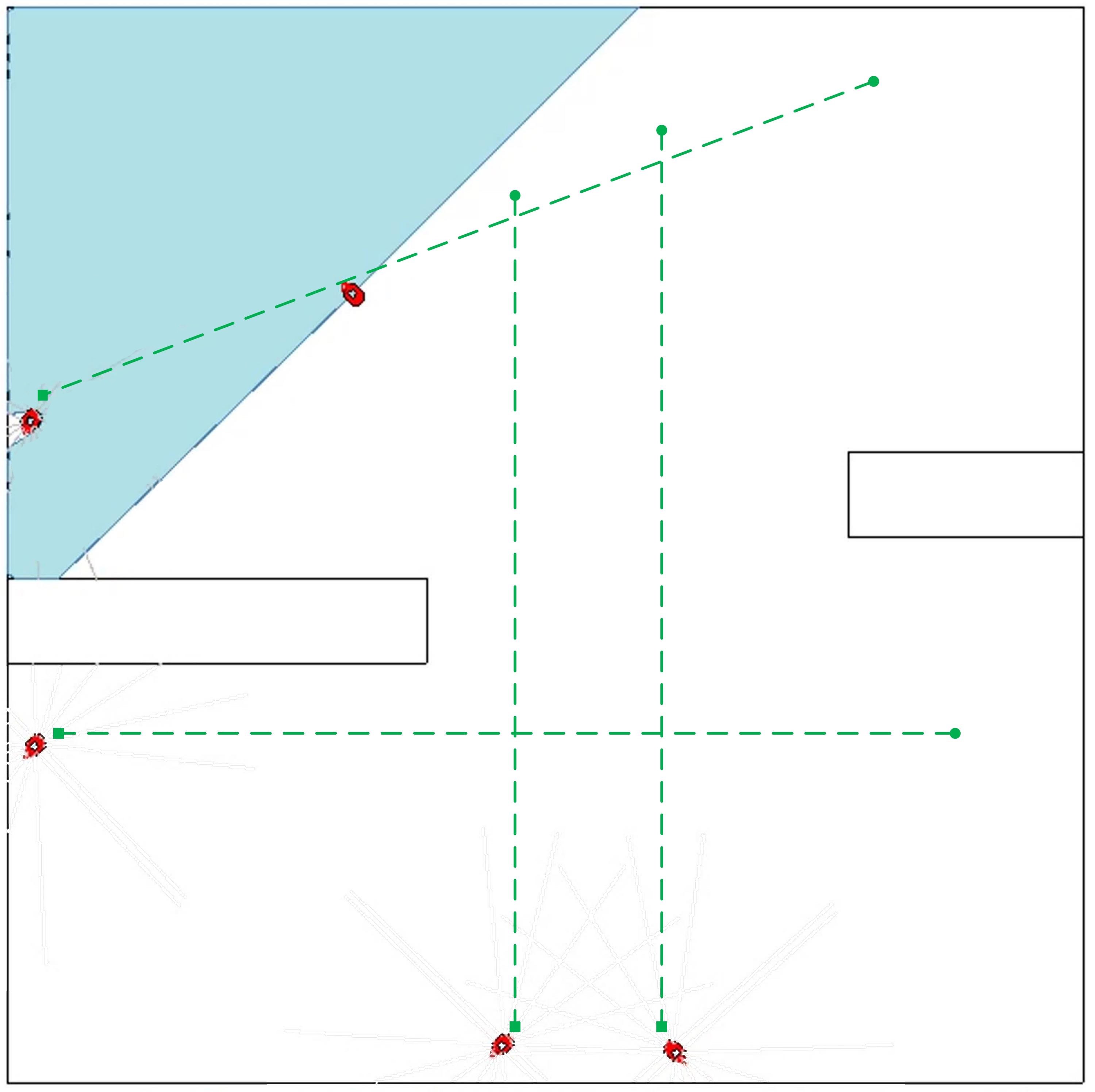} & \includegraphics[width=5.2cm,trim={12 4 8 105},clip]{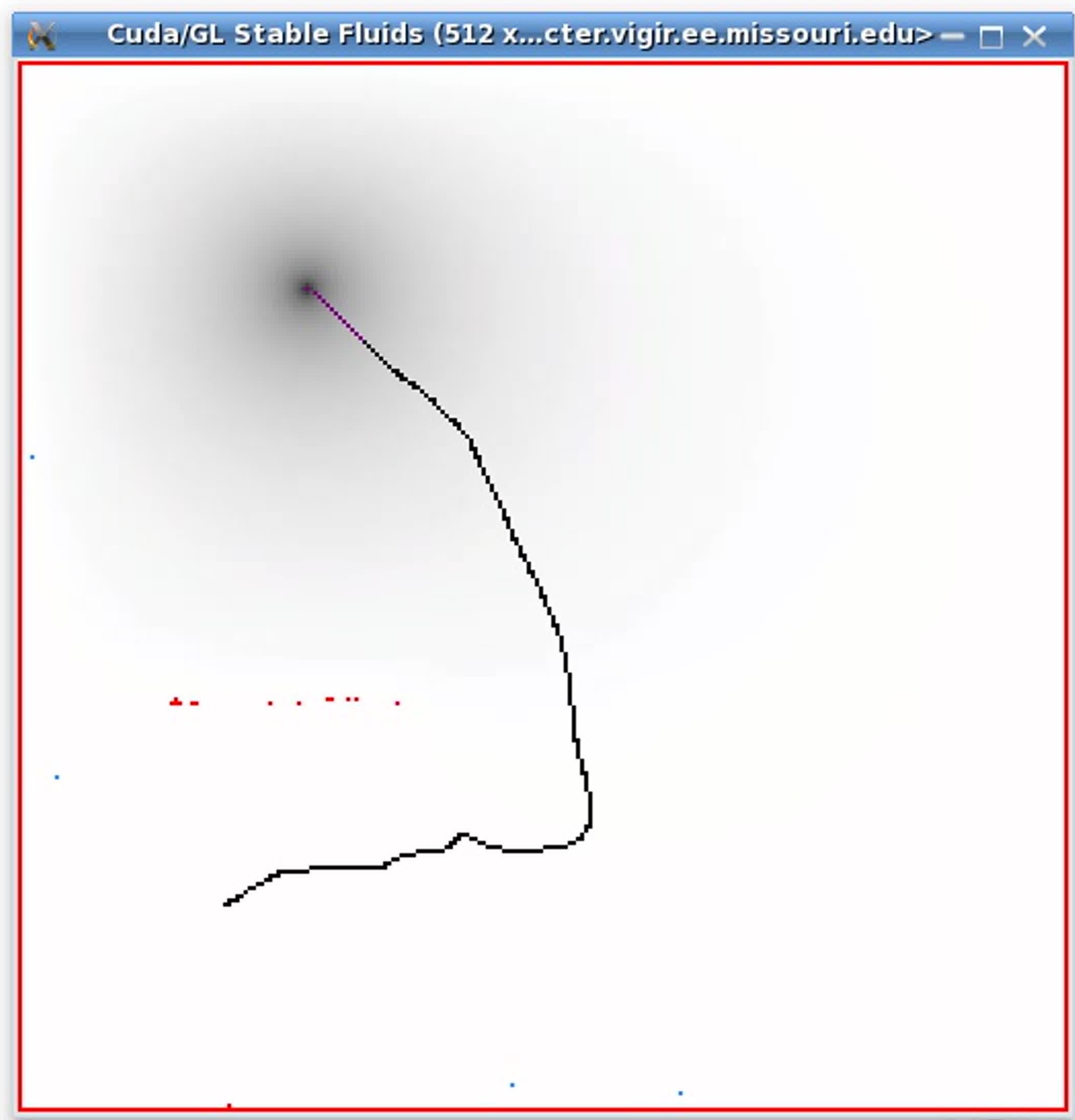}\tabularnewline
\multicolumn{2}{c}{{\small{}(c) 4 moving obstacles, map $\# $ 2.}}\tabularnewline
\end{tabular}
\caption{\label{fig:pp_results_1}Path planning results for 1, 2 and 4 moving
obstacles (Left: actual environment, Right: program's internal representation).}
\end{figure}

\begin{figure}
\centering
\begin{tabular}{cc}
\includegraphics[width=5.2cm]{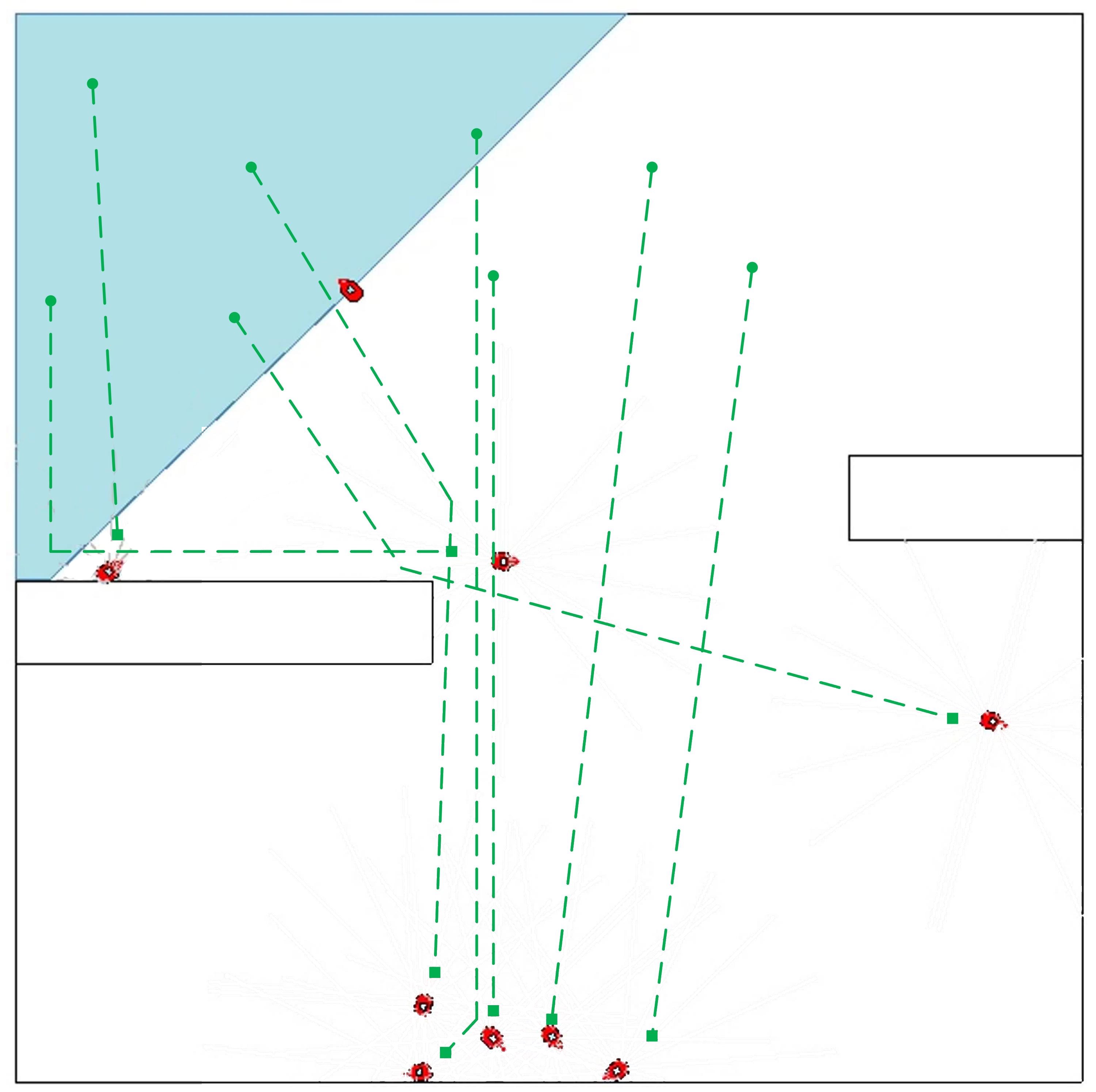} & \includegraphics[width=5.2cm,trim={10 4 10 90},clip]{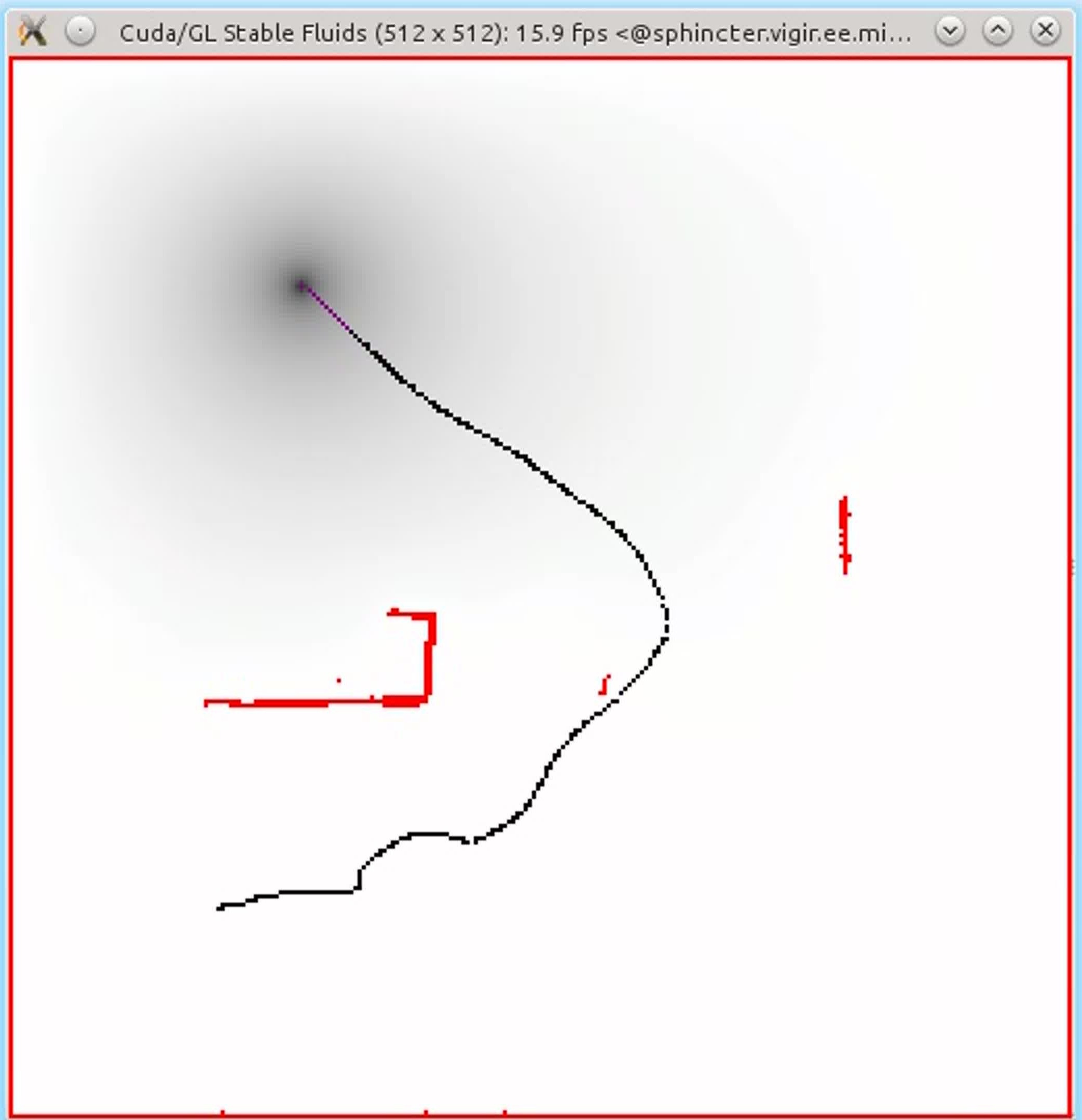}\tabularnewline
\multicolumn{2}{c}{{\small{}(a) 8 moving obstacles, map $\# $ 2.}}\tabularnewline
\includegraphics[width=5.2cm]{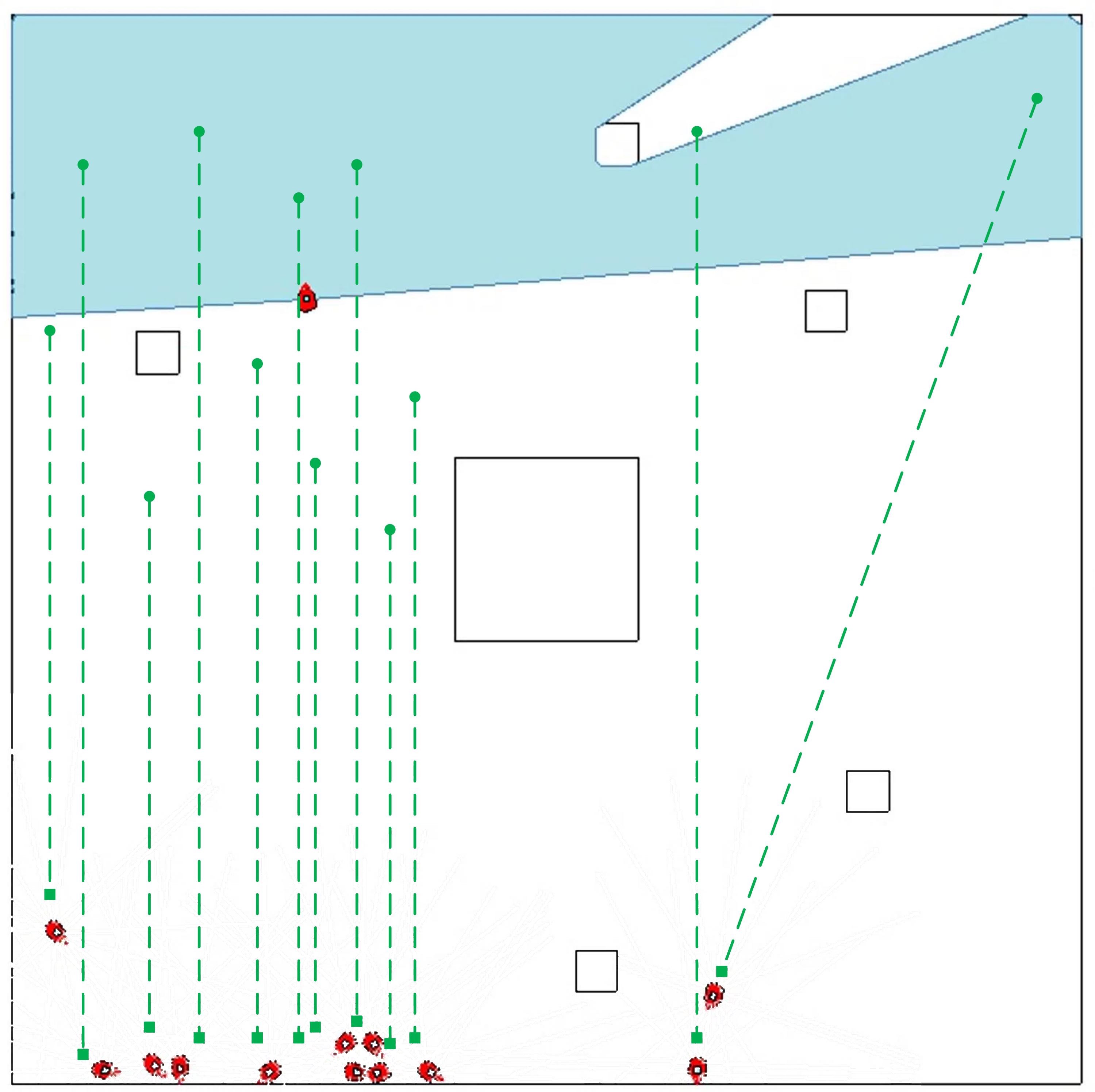} & \includegraphics[width=5.2cm,trim={10 4 10 95},clip]{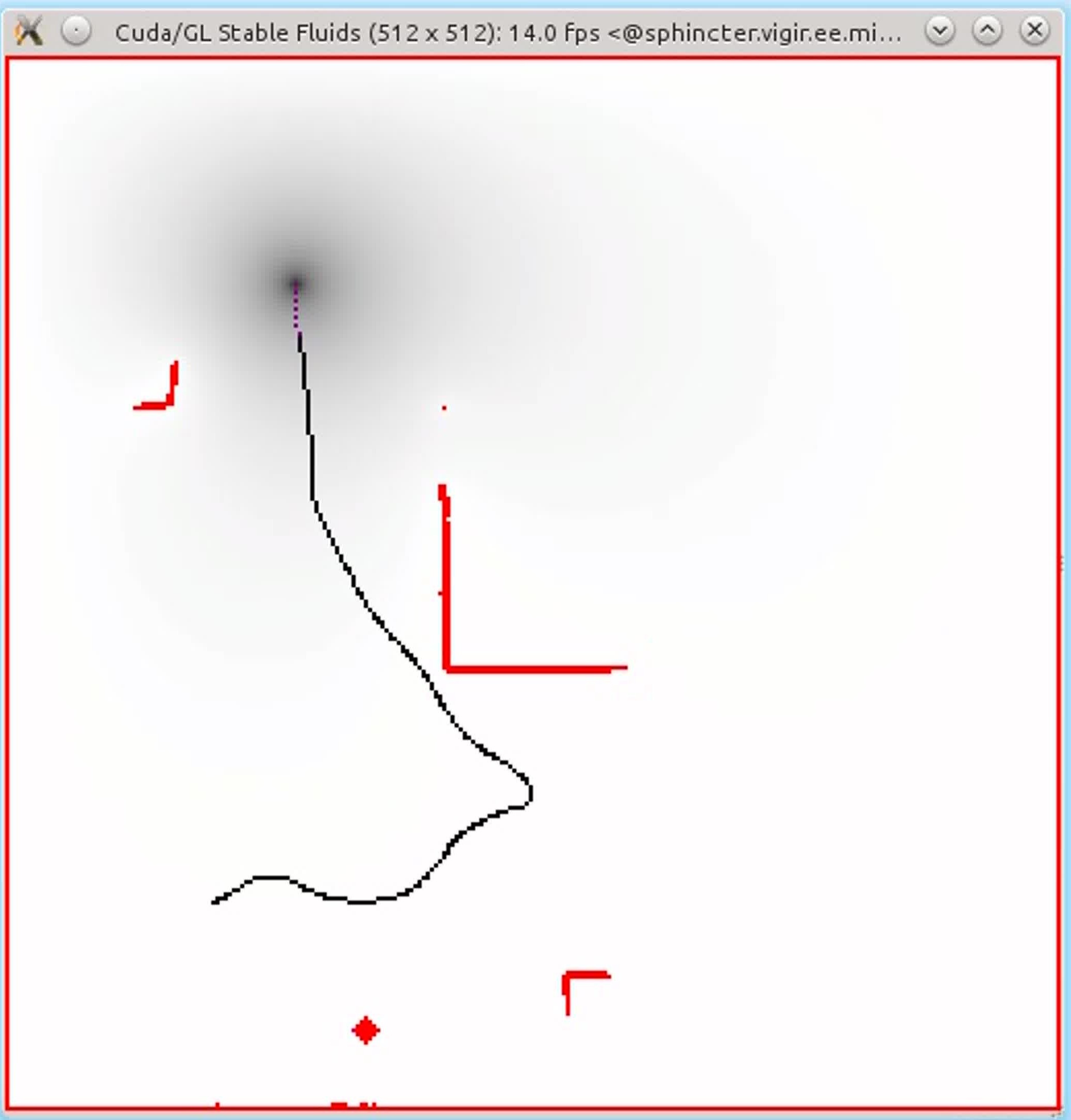}\tabularnewline
\multicolumn{2}{c}{{\small{}(b) 12 moving obstacles, map $\# $ 1.}}\tabularnewline
\includegraphics[width=5.2cm]{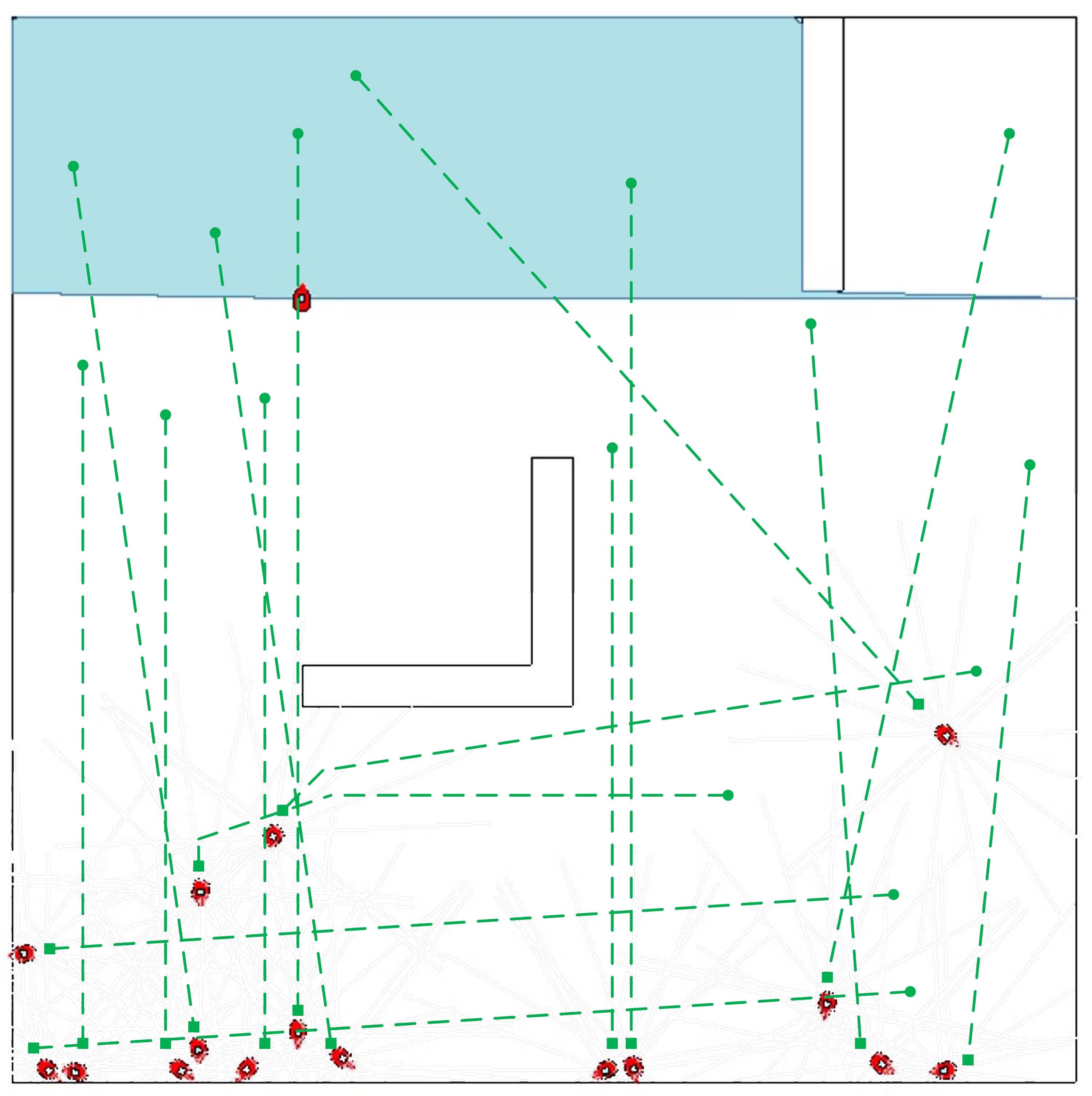} & \includegraphics[width=5.2cm,trim={10 4 10 90},clip]{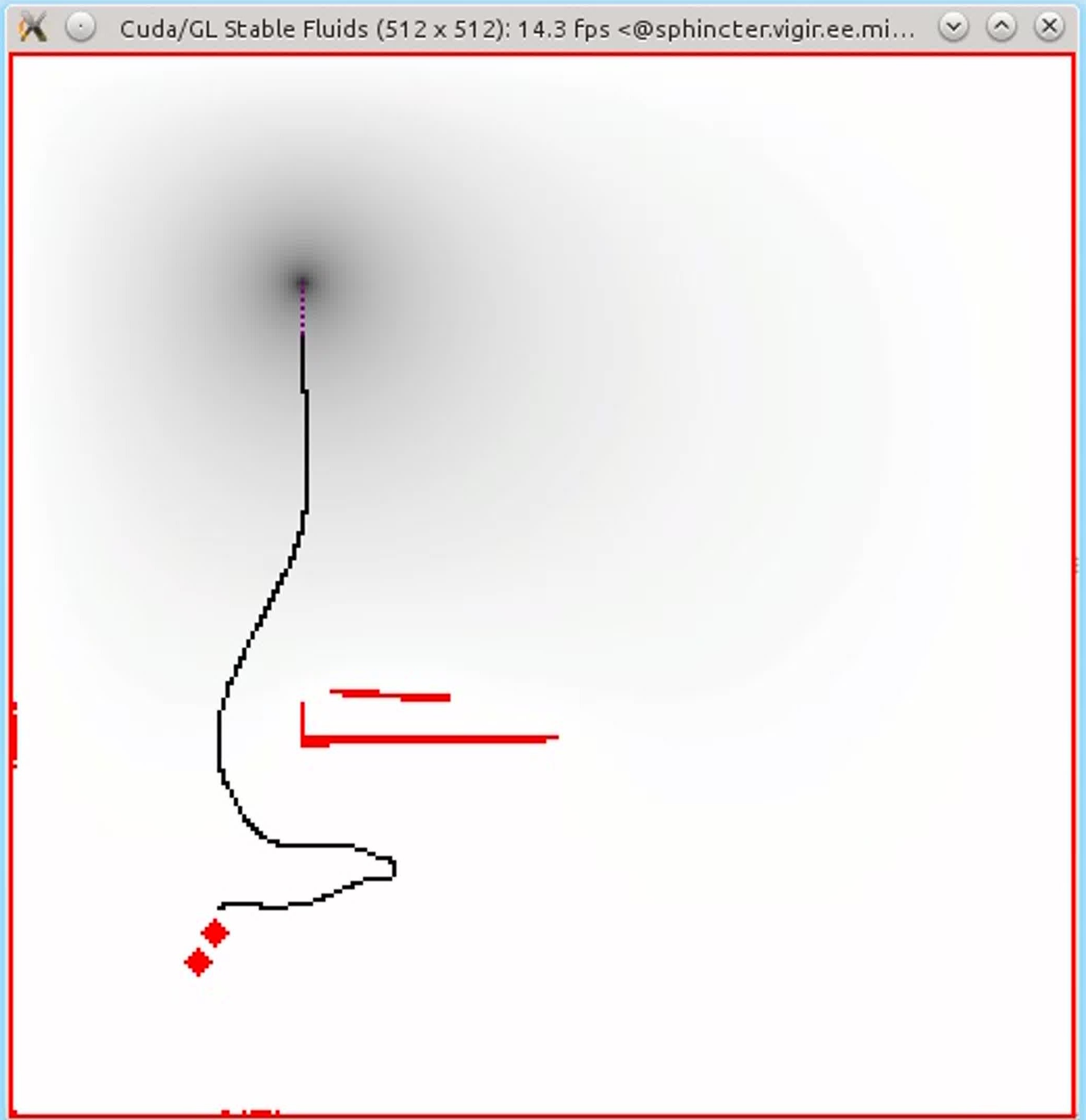}\tabularnewline
\multicolumn{2}{c}{{\small{}(c) 16 moving obstacles, map $\# $ 3.}}\tabularnewline
\end{tabular}
\caption{\label{fig:pp_results_2}Path Planning results for 8, 12 and 16 moving obstacles (Left: actual environment, Right: program's internal representation).}
\end{figure}

\begin{figure}[h]
\begin{tabular}{cc}
\includegraphics[width=0.35\columnwidth]{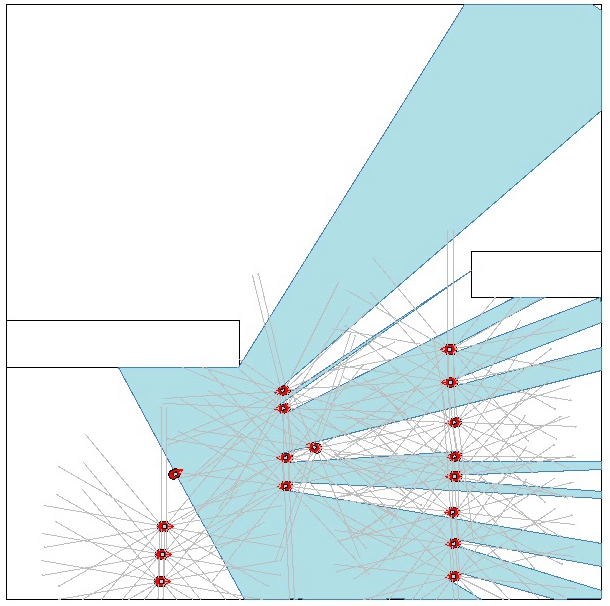} &
\includegraphics[width=0.34\columnwidth,trim={2 0 2 20},clip]{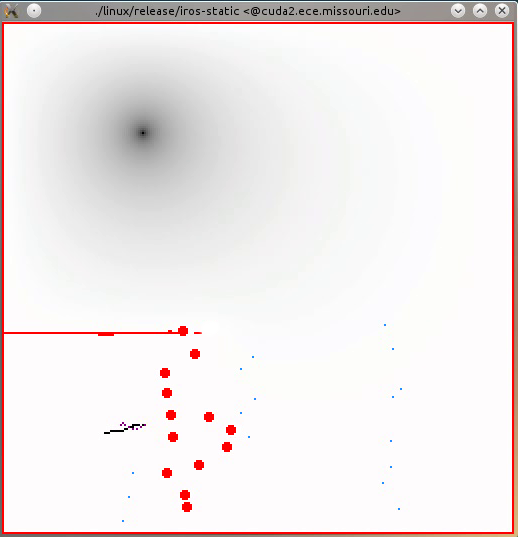}
\tabularnewline
{\small (a)} & {\small (b)}\tabularnewline
\end{tabular}
\vspace{10pt}
\begin{tabular}{cc}
\includegraphics[width=0.35\columnwidth]{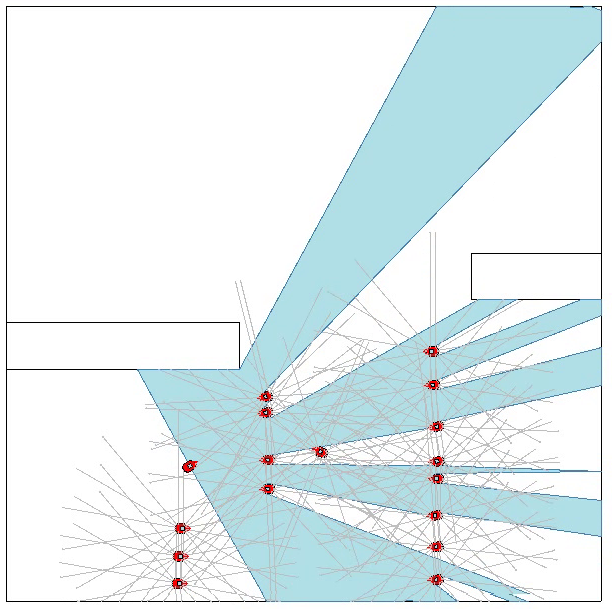} & 
\includegraphics[width=0.34\columnwidth,trim={2 0 2 18},clip]{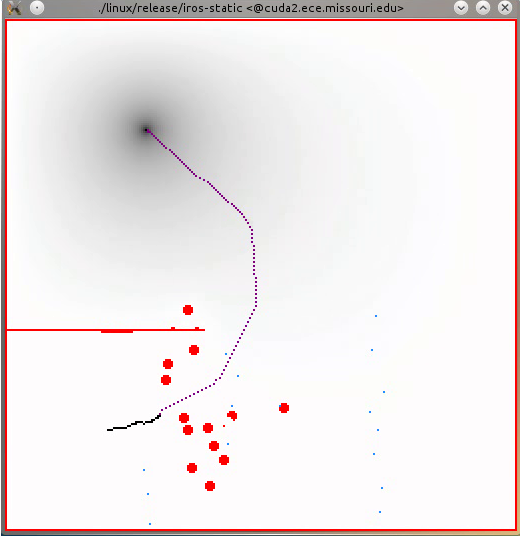}
\tabularnewline
\small (c) & \small (d)\tabularnewline
\end{tabular}
\vspace{-0.1in}
\centering{}\caption{\label{fig:nopass}(a, b) A scene when there is no planned path at the moment. (c, d) The algorithm can plan the path quickly.}
\end{figure}

For Figures \ref{fig:pp_results_1}-\ref{fig:nopass}, the
left window is the output from MobileSim, while the right window was created by the main program using OpenGL/CUDA to depict the internal representation of the sensed information form the environment, including the harmonic potentials, estimated position of obstacles, etc.  Also in the left window
of Figures \ref{fig:pp_results_1} and \ref{fig:pp_results_2}, the dashed green lines illustrate the movement of obstacles, and the solid black lines in the right window show the path traversed by the mobile robot. The red squares
show the estimated position of the moving obstacles in future, which are updated
in real time, while the red lines demonstrate the static obstacles sensed by the laser sensor of the robot. The reader should keep in mind that these figures display simply a snapshot of the environment and the final traversed path at the end of the tests. However, the moving obstacles may have caused the path to shift, which cannot be easily seen by these snapshots. 
Simulation videos of the the performance of the proposed path planning algorithm can be found in the accompanying video for the paper, available at
\url{http://vigir.missouri.edu/Time-Warp-Grid/}.

Figures \ref{fig:pp_results_1} and \ref{fig:pp_results_2} show some
of these scenarios for different number of moving obstacles. 
The robot was able to find a smooth and short path while avoiding collisions in the majority of the scenarios (Table \ref{tab:success}).


Our first evaluations were exactly regarding these three metrics:
success in avoiding collisions; length; and smoothness of the path.
The proposed approach was tested over 400 times in different environments,
all with size of $25.6 \times 25.6$ meters. The percentage of success for different
numbers of moving obstacles is presented in Table \ref{tab:success}.
A very strict criterion was used to determine success or failure:
if the robot collided with any obstacle before reaching the destination,
or if any moving obstacle was forced to change direction because of
the robot, the test was considered a failed attempt. It is important
to mention that with the proposed algorithm, the robot can never stop
and must always move with a constant velocity. 

Besides success rate, the algorithm was also tested for shortness
of the path found. Table \ref{tab:length} reports the average length of the
path for different maps and different number of obstacles. It would
be reasonable to expect the length of the path to grow with the number
of moving obstacles in the environment. However, this is not always
the case. Sometimes, random movement of obstacles cause the path to
change drastically and the final length increases even for a small
number of obstacles.

The last property of the path to be evaluated was regarding its smoothness.
Here, it should be mentioned that a path is usually
regarded as smooth if it does not intersect itself and its tangent
at each point varies continuously \cite{TayMan83}. Based on this definition, it can
be stated that all paths found by the algorithm were smooth, as it
is shown by Figures \ref{fig:pp_results_1} and \ref{fig:pp_results_2}. 

\begin{figure}[H]
    \centering
    \begin{tabular}{c}
    \includegraphics[width=0.7\columnwidth,trim={88 4 100 20},clip]{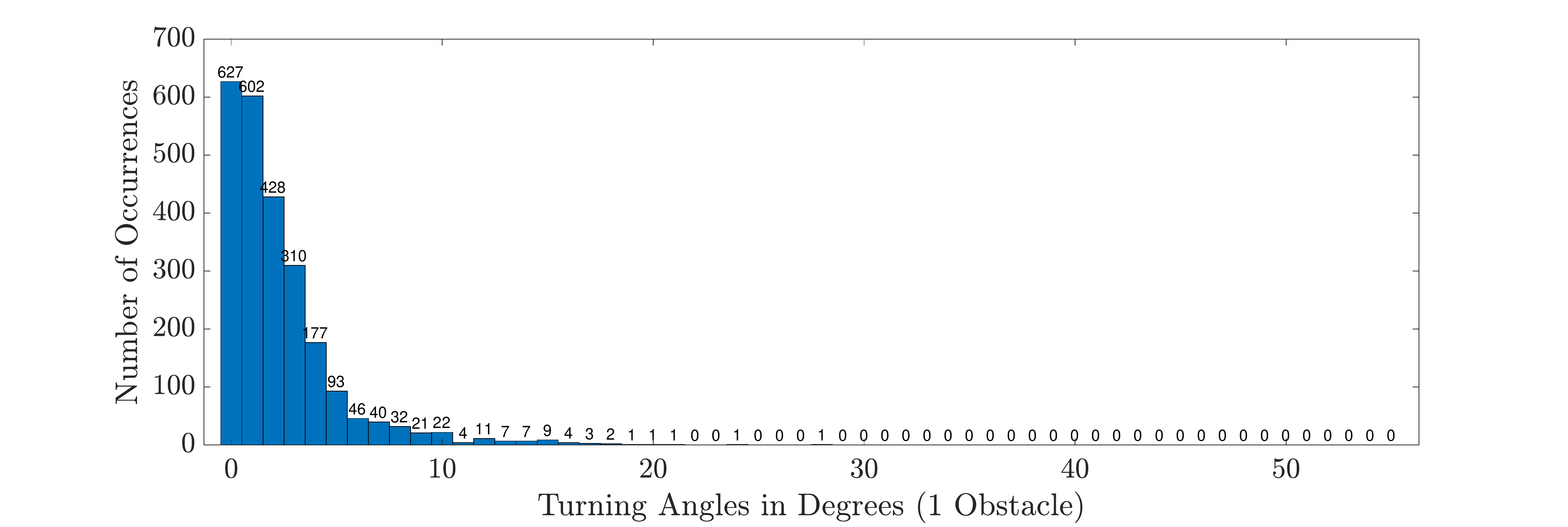} \tabularnewline
    \includegraphics[width=0.7\columnwidth,trim={88 4 100 20},clip]{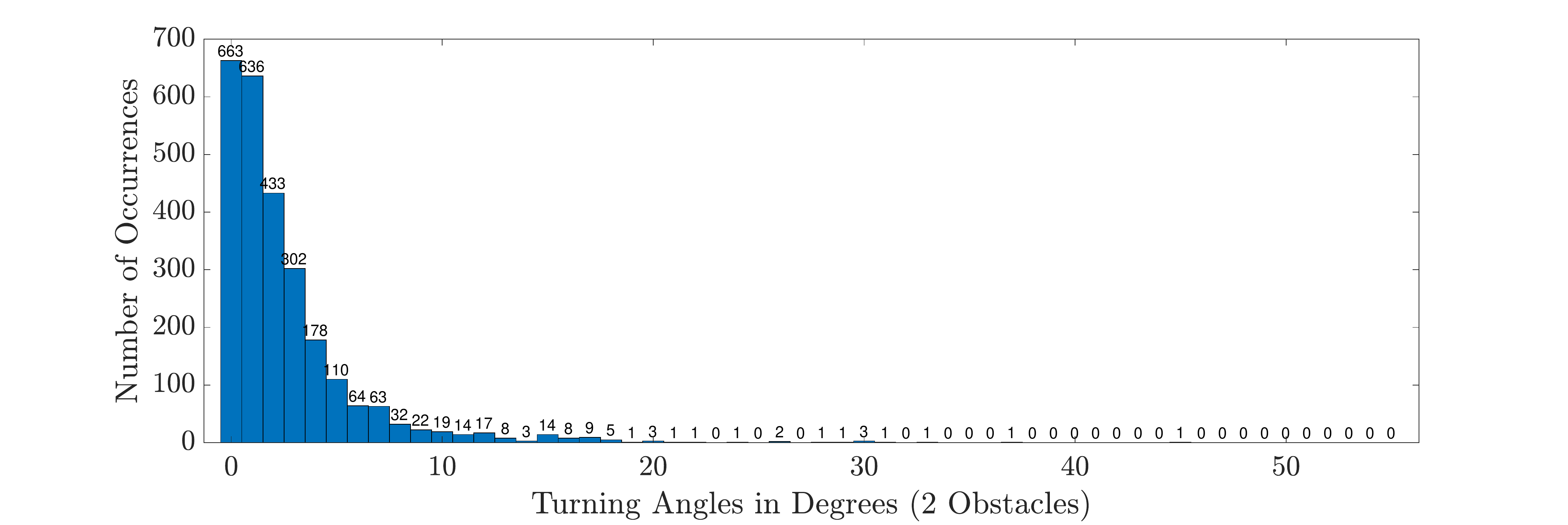} \tabularnewline
    \includegraphics[width=0.7\columnwidth,trim={88 4 100 20},clip]{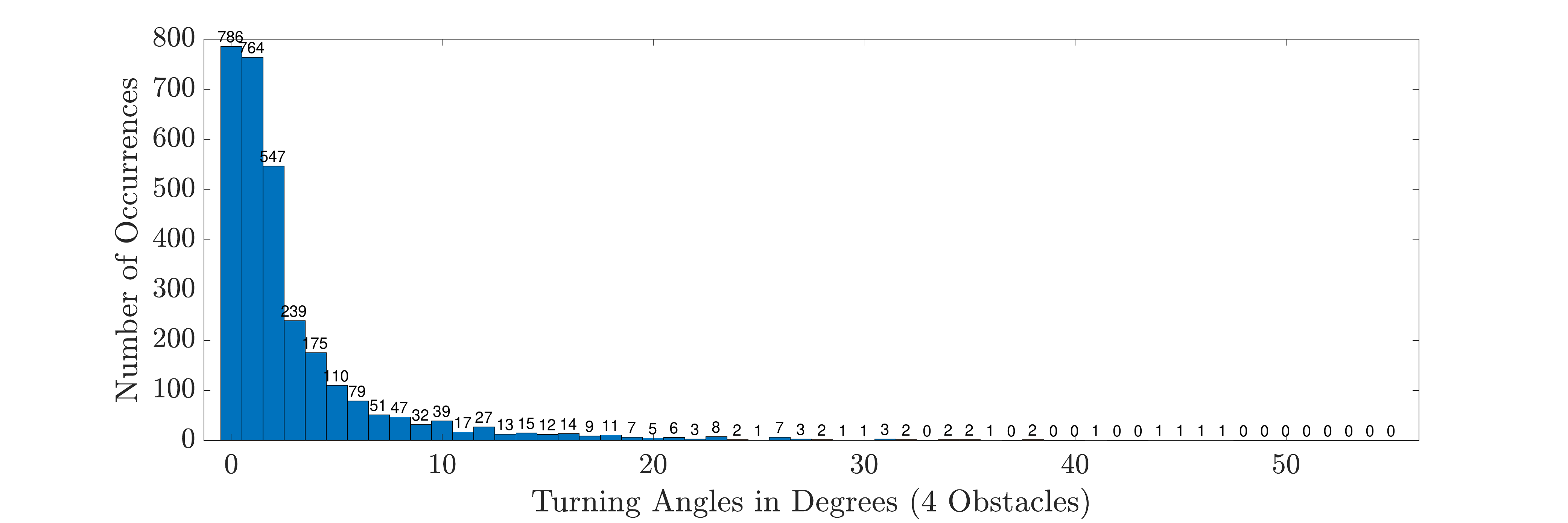} \tabularnewline
    \includegraphics[width=0.7\columnwidth,trim={88 4 100 20},clip]{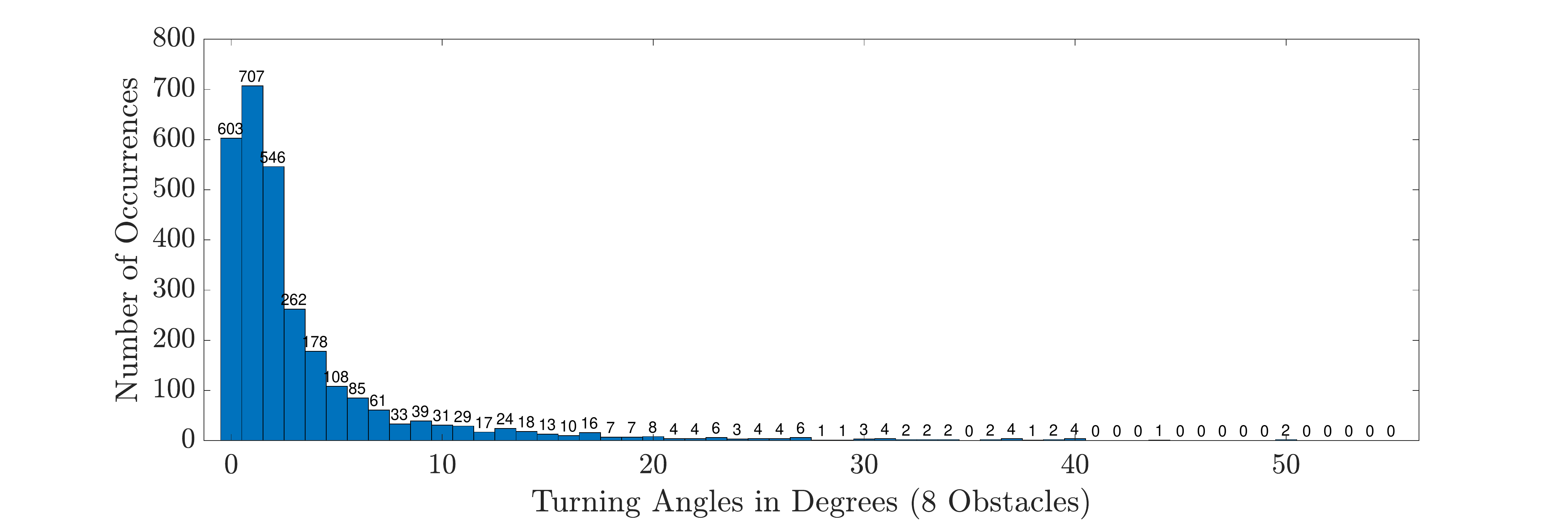} \tabularnewline
    \includegraphics[width=0.7\columnwidth,trim={88 4 100 20},clip]{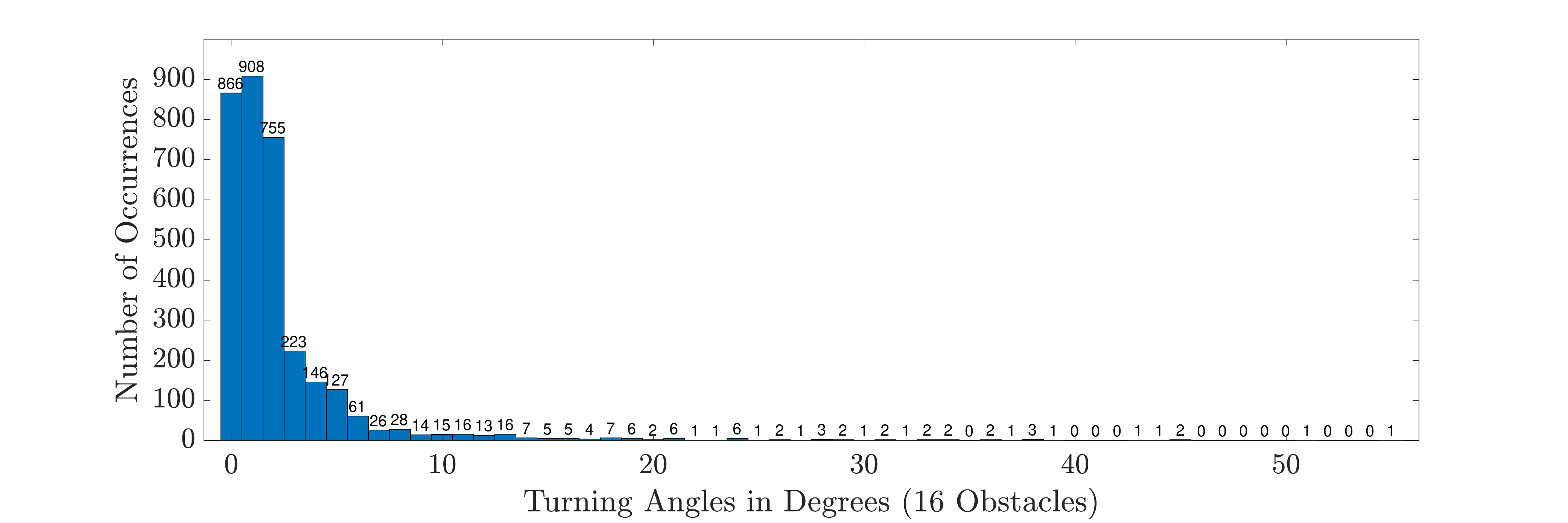} \tabularnewline
    \end{tabular}
    \caption{Distribution of turning angles (in degree) for 20 repetitions of the algorithm for 1, 2, 4, 8, and 16 moving objects.}
    \label{fig:angles}
\end{figure}

However, due to the target application of the proposed algorithm for
power wheelchairs, we also evaluated the algorithm for large turning
angles. So, in order to further quantify the smoothness of the optimized
path in a dynamic environment, we computed the histogram of turning
angles performed during navigation. Figure \ref{fig:angles}
shows the histograms for different numbers of moving obstacles. As
these plots indicate, the turning angles are mostly concentrated
on the small angular values. Moreover, even though increasing the
number of moving obstacles leads to larger turning angles, these are
very rare and likely due to some ``deadlocked'' situations.

\begin{figure}[!h]
\centering{}\includegraphics[width=0.35\columnwidth,trim={2 0 2 72},clip]{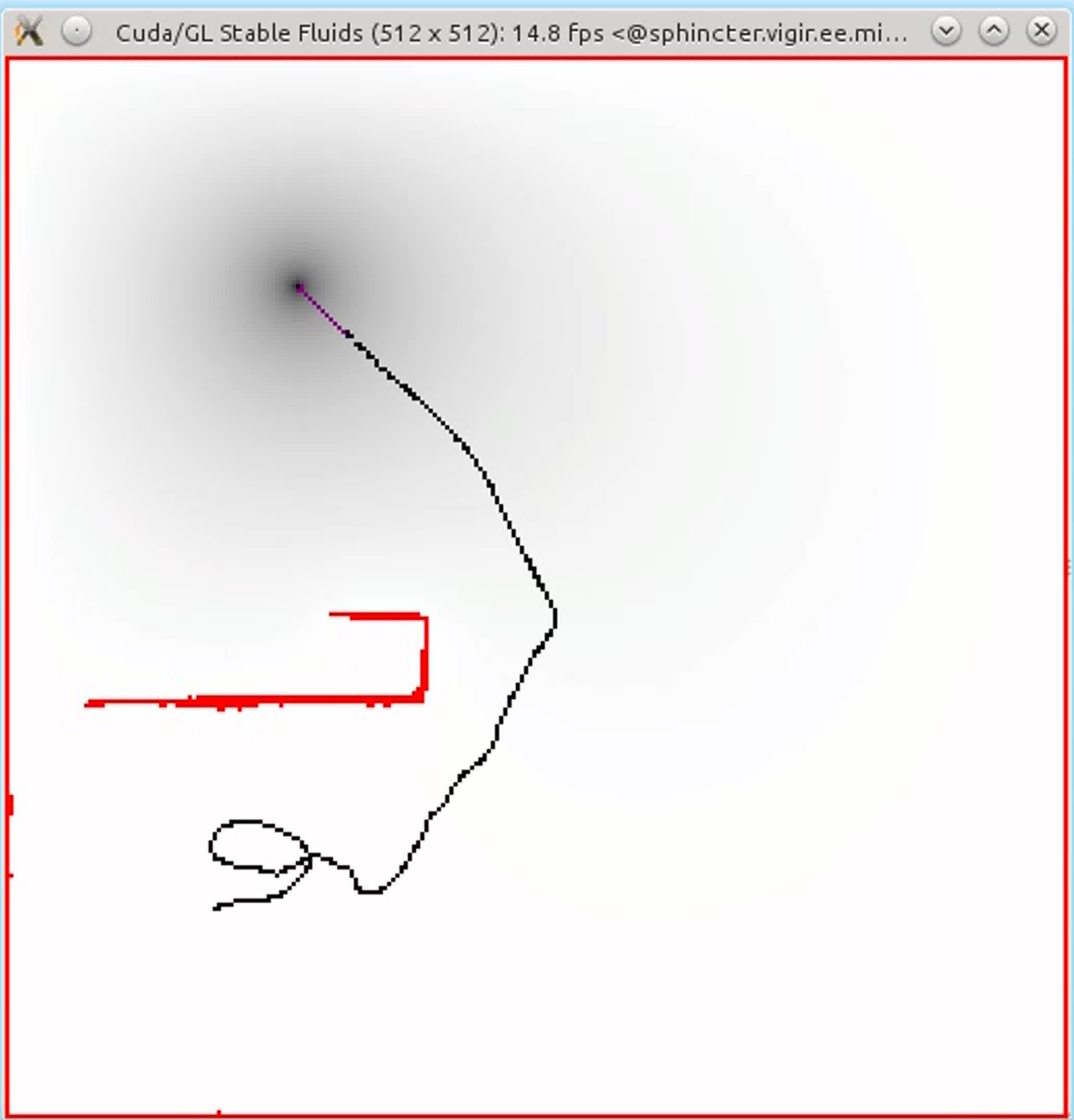}
\caption{{\small{}\label{fig:nopass3}}A loop in the path caused by a blocked
situation.}
\vspace{-0.1in}
\end{figure}

Indeed, in scenarios with a very large number of moving obstacles,
it might happen that these obstacles completely block the path of
the robot towards the goal. One such scenario is presented in Figure
\ref{fig:nopass} (a)-(b). However, even in those extreme cases, after a
while, the robot is able to re-trace a path towards the goal as the
obstacles move again away from the robot path (Figure \ref{fig:nopass} (c)-(d)).

\begin{table}[H]
\caption{{\small \label{tab:success}Percentage of success in path planning
for different scenarios (out of 400 simulations).}}
\centering
\begin{tabular}{ccccccc}
\toprule
{\scriptsize{}Number of moving obstacles} & {\scriptsize{}1} & {\scriptsize{}2} & {\scriptsize{}4} & {\scriptsize{}8} & {\scriptsize{}12} & {\scriptsize{}16}\tabularnewline
\midrule 
{\scriptsize{}Percentage of success} & {\scriptsize{}95.0 \%} & {\scriptsize{}95.0 \%} & {\scriptsize{}92.5 \%} & {\scriptsize{}87.5 \%} & {\scriptsize{}80.0 \%} & {\scriptsize{}75.0 \%}\tabularnewline
\bottomrule
\end{tabular}
\end{table}

\begin{table}[H]
\vspace{-0.17in}
\caption{{\small \label{tab:length}Average length of the traversed path (cm).}}
\centering
\begin{tabular}{ccccccc}
\toprule
{\scriptsize{}Number of moving obstacles} & {\scriptsize{}1} & {\scriptsize{}2} & {\scriptsize{}4} & {\scriptsize{}8} & {\scriptsize{}12} & {\scriptsize{}16}\tabularnewline
\midrule 
{\scriptsize{}Map \#1} & {\scriptsize{}4610} & {\scriptsize{}5400} & {\scriptsize{}5270} & {\scriptsize{}4970} & {\scriptsize{}5955} & {\scriptsize{}7085}\tabularnewline
{\scriptsize{}Map \#2} & {\scriptsize{}6705} & {\scriptsize{}6955} & {\scriptsize{}7665} & {\scriptsize{}6475} & {\scriptsize{}6415} & {\scriptsize{}6965}\tabularnewline
{\scriptsize{}Map \#3} & {\scriptsize{}4850} & {\scriptsize{}4680} & {\scriptsize{}4725} & {\scriptsize{}6800} & {\scriptsize{}7380} & {\scriptsize{}8215}\tabularnewline
{\scriptsize{}Map \#4} & {\scriptsize{}4870} & {\scriptsize{}4875} & {\scriptsize{}5045} & {\scriptsize{}5090} & {\scriptsize{}5660} & {\scriptsize{}6885}\tabularnewline
\bottomrule
\end{tabular}
\end{table}

Here, we should mention that due to the limitation of the algorithm
of not controlling the velocity of the robot, if the blocking situation
persists for long enough, the robot has to turn around until it finds
a new path, causing loops and sharp turns in the path (Figure \ref{fig:nopass3}). These deadlock situations occur when the moving obstacles are aligned in a linear formation in a very dense area, or when multiple moving obstacles try to negotiate a narrow passage, e.g. a doorway. Deadlock situations can be avoided by adapting the algorithm to control the velocity of the mobile robot, which is beyond the scope of this work and will be the subject of future work.

\section{Conclusion}

This paper presented a unique smooth path planning approach for dynamic
and unknown environments for mobile robots. The algorithm
relies on: harmonic potential fields to build a path; rubber band
model to smoothen the path; and an integration of Kalman Filter and
a new idea of Time-Warped Grid to estimate the position of moving
obstacles and avoid collisions. The concept of time-warped grid reduces
and simplifies many processes by eliminating the need to take the
directions of movement and the absolute value of the distance between
the robot and the moving obstacles into consideration during the estimation process.
The proposed method was tested exhaustively using several simulation
scenarios for the Pioneer P3-DX robot. The implementation of the algorithm
was carried out using C/C++ and CUDA programming for real-time performance.
As the simulations demonstrated, our approach is robust and is able to find the optimum path -- i.e in terms of smoothness, distance, and collision-free -- either in static or dynamic environments, with very high degree of success even with a very large number of moving obstacles.

\section{Future Work}

In the course of this research project we have identified many directions
for future work. Some are particular to our proposed approach, and
some are broader in scope. The vision-based landmark localization
system could be replaced by an algorithm to distinguish dynamic and
static obstacles using only a laser sensor, which may result in a
simpler system. Furthermore, using embedded GPUs like NVIDIA Jetson enables us to run the algorithm on the robot itself, rather than having an external server and consequent communications.

The dynamic obstacles in this work are assumed to have linear motion (they turn only when they get close to walls and other obstacles) and move with a constant velocity, resulting in a simple state matrix and dynamics used by the Kalman filter algorithm. In case of more complicated behavior for the moving obstacles, the Kalman filter should be built using the associated dynamic equations. Machine learning techniques such as neural networks can also be employed to learn the model and estimate the future positions of the moving obstacles and collision points.

Finally, an algorithm that could adopt the velocity of the mobile robot
-- or even stop it -- should be investigated to solve the rare,
but existing cases of momentary deadlocks.

\printbibliography

\end{document}